\documentclass[10pt,twocolumn,letterpaper]{article}

\usepackage{iccv}      

\usepackage[utf8]{inputenc} 
\usepackage[T1]{fontenc}    
\usepackage{url}            
\usepackage{booktabs}       
\usepackage{amsfonts} 
\usepackage{comment}
\usepackage{nicefrac}       
\usepackage{microtype}      
\usepackage{xcolor}         
\usepackage{amsmath,bm}
\usepackage{float}
\usepackage{multirow, graphicx}
\usepackage{graphicx}
\usepackage{tabularx, makecell, multirow} 
\usepackage{wrapfig}
\usepackage{amsmath, amssymb, amsthm, geometry, mathtools}
\usepackage{graphicx}

\usepackage{amsfonts}

\usepackage{subcaption}    

\usepackage{amsmath,amssymb}

\usepackage[dvipsnames]{xcolor}
\usepackage{xcolor,colortbl}
\usepackage{multirow, makecell}
\definecolor{Gray}{gray}{0.50}
\newcolumntype{g}{>{\columncolor{Gray}}c}
\definecolor{ffe1da}{RGB}{255,225,218}
\definecolor{F7E0D5}{RGB}{247,224,213}
\definecolor{darkF7E0D5}{RGB}{209,154,128}
\colorlet{Light}{White!0!F7E0D5}

\colorlet{tabfirst}{Green!25}
\definecolor{tabthird}{rgb}{1, 0.85, 0.7}
\definecolor{tabsecond}{rgb}{1, 0.96, 0.7}

\definecolor{tabfirst}{rgb}{0.79, 0.92, 1.0} 
\definecolor{tabsecond}{rgb}{1.0, 0.86, 0.86} 
\definecolor{tabthird}{rgb}{0.9, 0.8, 1.0} 

\definecolor{cvprblue}{rgb}{0.21,0.49,0.74}
\definecolor{mypurple}{rgb}{0.4549,0.145,0.99}

\usepackage{tikz}
\usepackage{xcolor} 
\usepackage{adjustbox}
\usepackage{amsmath}

\usepackage{natbib}
\usepackage{siunitx}

\usepackage{amsmath}
\usepackage{algorithm}
\usepackage{algorithmicx}
\usepackage{algpseudocode}
\usepackage{amsmath} 
\newcommand{\mat}[1]{\mathbf{#1}} 
\usetikzlibrary{shapes.geometric, shadows} 
\graphicspath{{figs/}}

\definecolor{iccvblue}{rgb}{0.21,0.49,0.74}
\definecolor{surfacecolor}{RGB}{255,145,145}
\definecolor{insidecolor}{RGB}{113,191,225}
\definecolor{outsidecolor}{RGB}{252,201,126}

\DeclareRobustCommand{\colorpentagram}[2]{
  \begin{tikzpicture}[scale=5] 
    \node[star, 
          star points=5, 
          fill=#1, 
          minimum size=0.5mm,    
          inner sep=0.1pt,         
          draw=black, 
          line width=0.1pt,      
          transform shape        
         ] {};
    \node[white, font=\fontsize{3}{3}\selectfont] at (0,0) {#2}; 
  \end{tikzpicture}%
}
\usepackage[pagebackref,breaklinks,colorlinks,allcolors=iccvblue]{hyperref}

\def\paperID{1584} 
\def\confName{ICCV}
\def\confYear{2025}

\title{Bridging 3D Anomaly Localization and Repair via \\ High-Quality Continuous Geometric Representation }
\author{
    $\text{Bozhong Zheng}^{1}\thanks{Equal contribution}$ \quad
    $\text{Jinye Gan}^{1\ast}$ \quad 
    $\text{Xiaohao Xu}^{2\dagger}$ \quad 
    $\text{Xintao Chen}^{1}$ \quad \\
    $\text{Wenqiao Li}^{1}$ \quad
    $\text{Xiaonan Huang}^{2}$\quad
    $\text{Na Ni}^{1}\thanks{Corresponding authors}$ \quad 
    $\text{Yingna Wu}^{1\dagger}$ \\
    $^{1}$ShanghaiTech University \quad $^{2}$University of Michigan, Ann Arbor   \\}

\begin{document}
\maketitle
\begin{abstract}

3D point cloud anomaly detection is essential for robust vision systems but is challenged by pose variations and complex geometric anomalies. Existing patch-based methods often suffer from geometric fidelity issues due to discrete voxelization or projection-based representations, limiting fine-grained anomaly localization.
We introduce Pose-Aware Signed Distance Field (PASDF), a novel framework that integrates 3D anomaly detection and repair by learning a continuous, pose-invariant shape representation. PASDF leverages a Pose Alignment Module for canonicalization and a SDF Network to dynamically incorporate pose, enabling implicit learning of high-fidelity anomaly repair templates from the continuous SDF. This facilitates precise pixel-level anomaly localization through an Anomaly-Aware Scoring Module.
Crucially, the continuous 3D representation in PASDF extends beyond detection, facilitating in-situ anomaly repair. Experiments on Real3D-AD and Anomaly-ShapeNet demonstrate state-of-the-art performance, achieving high object-level AUROC scores of 80.2\% and 90.0\%, respectively. These results highlight the effectiveness of continuous geometric representations in advancing 3D anomaly detection and facilitating  practical anomaly region repair. 
The code is available at \url{https://github.com/ZZZBBBZZZ/PASDF} to support further research.
\end{abstract}    

\section{Introduction}
\label{sec:intro}
Automated 3D anomaly detection is rapidly becoming vital across computer vision applications\cite{cao2024survey,liu2024deep}, from quality control in manufacturing to robust robotic manipulation. In scenarios like component inspection, even subtle anomalies—a missing feature, deformation, or shape irregularity—can compromise entire assemblies.  Robust 3D anomaly detection techniques, focused on individual objects in isolation, are thus essential\cite{Real3d-AD}. 

\begin{figure}[t!]
    \centering \setlength{\abovecaptionskip}{0.1cm}
    \includegraphics[width=1\linewidth]{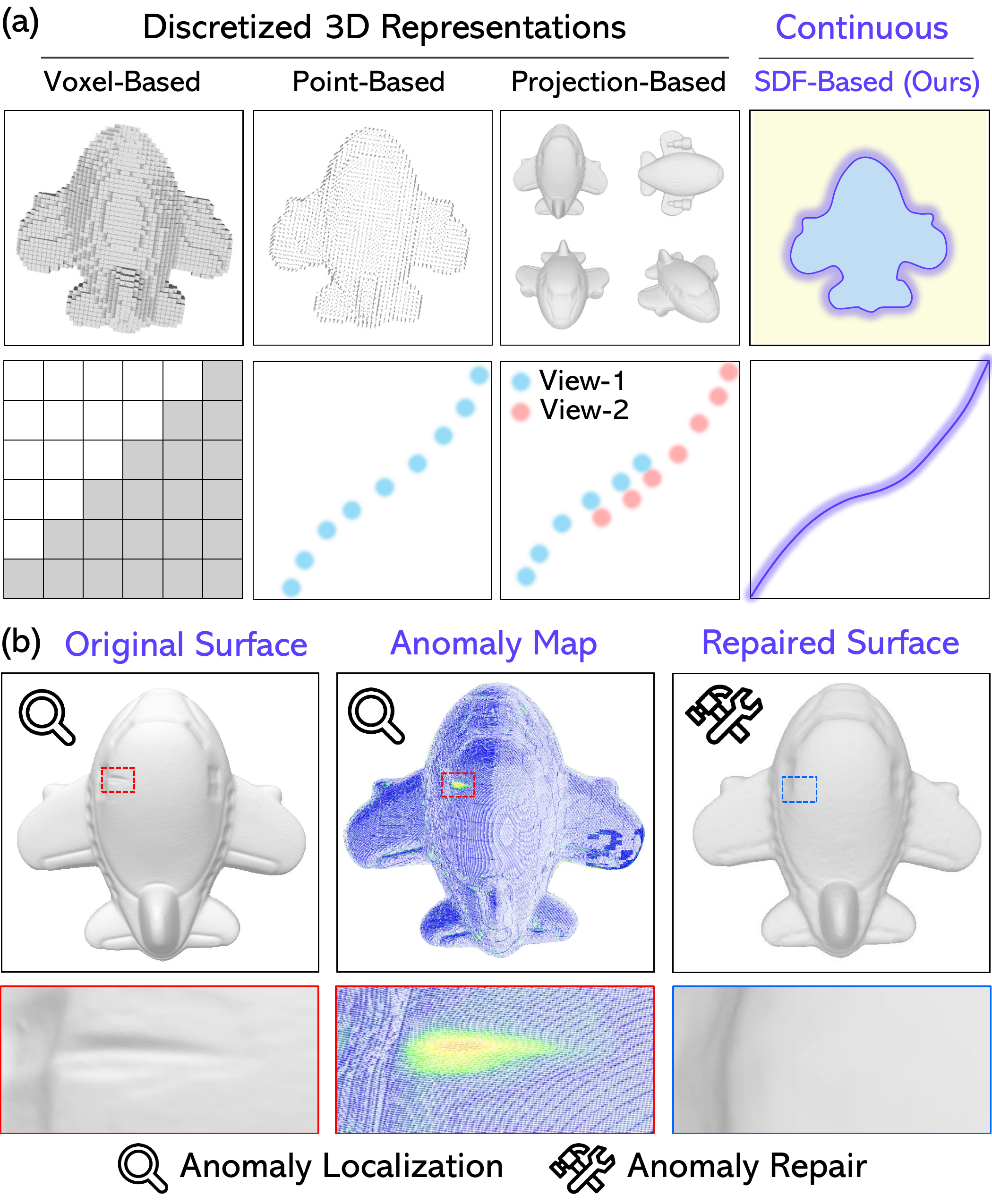}
\caption{(\textbf{a}) 3D representation comparison for anomaly detection: voxel-based, point-based, projection-based, and our approach with signed distance function (SDF). (\textbf{b}) PASDF's continuous geometric representation enables both high-precision anomaly localization and effective anomaly surface repair.}
    \label{fig:teaser_figure}\vspace{-2mm}
\end{figure}

Existing 3D anomaly detection methods often rely on \textbf{discretized 3D representations}—voxel\cite{Bergmann_2022_mvtec_3dad}, point\cite{Real3d-AD,Group3AD}, or projection-based\cite{cao2023complementary,cheng2024towards}. While computationally efficient, these representations inherently sacrifice fine-grained geometric detail and introduce quantization artifacts (see Fig.~\ref{fig:teaser_figure}a), critical limitations for object-level anomaly localization. Subtle shape deviations, such as warped surfaces or incorrectly sized features, can be obscured by voxel grid limitations\cite{horwitz2023back}. Furthermore, voxel-based methods face cubic memory scaling with resolution, hindering their applicability to complex, high-resolution 3D objects. Point-based methods, though avoiding volumetric discretization, struggle with inconsistent density and incomplete surface coverage due to sparse sampling\cite{yang20203dssd,qian2022pointnext}, while projection-based approaches suffer from information loss\cite{shi2019pointrcnn,guo2020deep} in occluded regions and viewpoint-dependent distortions.  These limitations motivate a novel approach capable of capturing global shape, robust to pose variations, and preserving fine geometric details without discretization, specifically for \textit{object-level} anomaly identification.

Crucially, in real-world applications like 3D printing and advanced manufacturing, anomaly detection is not the end goal—\textbf{anomaly repair is equally vital}\cite{singh2023toward}. Identifying defects is only the first step; in-situ defect correction necessitates accurate reconstruction of the normal object as a repair template.  Conventional memory-based anomaly detection\cite{Real3d-AD,Group3AD,cao2023complementary}, relying on indirect feature mappings, struggles to guide repair due to a lack of explicit shape reconstruction and geometric fidelity. Reconstruction-based methods, like IMRNet\cite{IMRNet} and R3D-AD\cite{R3D-AD}, offer a more promising avenue by attempting to restore anomalous point clouds. However, their reliance on \textbf{discrete point-based representations inherently limits their ability to generate continuous, high-fidelity repair templates}, resulting in incomplete reconstructions and loss of fine details, insufficient for high-precision repair tasks.

To overcome these limitations and \textbf{unify 3D anomaly detection and repair} (see Fig.~\ref{fig:teaser_figure}b), we introduce {Pose-Aware Signed Distance Function (PASDF)}, a novel unsupervised framework leveraging the representational power of Signed Distance Functions (SDFs). SDFs offer a smooth, continuous shape representation, inherently capturing fine-grained geometric details essential for anomaly identification and repair template generation. PASDF's key innovation is the explicit disentanglement of pose and shape through a carefully designed architecture. We propose a {Pose-wise Alignment Module (PAM)} that robustly registers input point clouds to a learned canonical coordinate frame \textit{before} implicit SDF network processing. This pose alignment enables the SDF network to focus on intrinsic shape variations, decoupling shape representation from pose artifacts and mitigating the confounding effects of arbitrary object poses. By learning a canonicalized SDF representation, PASDF effectively captures global geometric context and detects subtle anomalies missed by local feature methods, while simultaneously generating a high-fidelity repair template.

Finally, we demonstrate PASDF's ability to visually identify anomalous objects with subtle shape deviations or missing components, even in arbitrary poses, such as detecting a missing mug handle or warped surface on a manufactured part—challenges for voxel-based methods due to quantization and pose variations. Rigorous evaluations on Real3D-AD and Anomaly-ShapeNet datasets showcase state-of-the-art \textbf{object-level} anomaly detection performance. Notably, PASDF achieves a 5\% O-AUROC improvement on Real3D-AD over prior works, highlighting its superior object-level anomaly identification.

\textbf{Our contributions are summarized as follows:}

\begin{itemize}
    \item    We propose PASDF, an unsupervised {object-level} 3D anomaly detection and repair framework via pose-aware SDFs, achieving high performance and robustness to pose variance of test cases. 
    \item We introduce PAM, a point cloud registration module enabling pose-invariant learning and eliminating discretization accuracy loss.
    \item We demonstrate SOTA object-level 3D anomaly detection on Real3D-AD and Anomaly-ShapeNet datasets, accompanied by compelling qualitative results and a capability for high-quality object template reconstruction which facilitates {in-situ anomaly repair}.
\end{itemize}

\begin{figure*}[t!]
  \centering
  \setlength{\abovecaptionskip}{0.3cm}
  \includegraphics[width=0.95\textwidth]{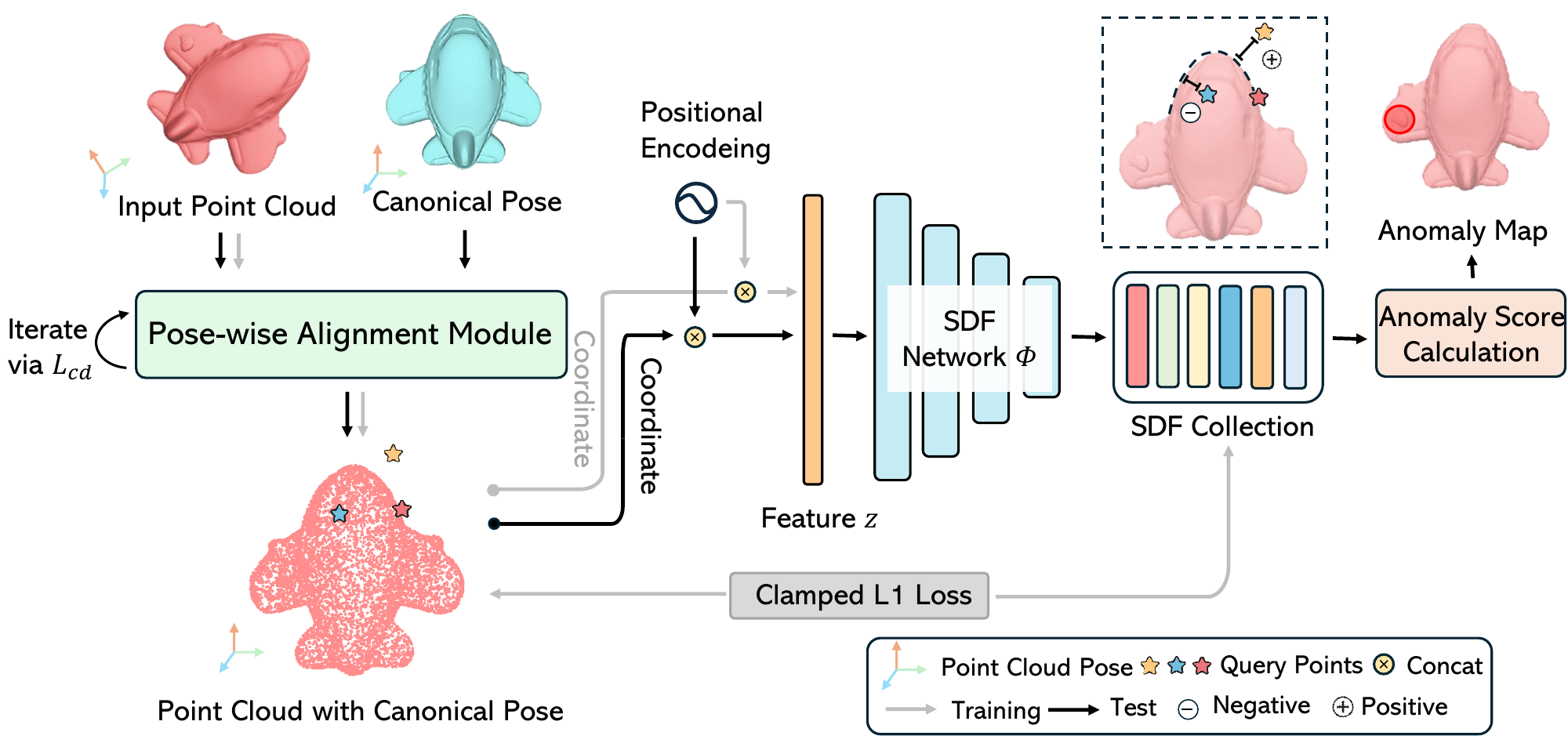}
  \caption{\textbf{Overview of \textit{Pose-Aware Signed Distance Function (PASDF)} for 3D Anomaly Detection.} A normal point cloud is selected as the Canonical Pose, and all training and test point clouds are aligned to it via the Pose-wise Alignment Module. Query points are categorized as surface (\colorpentagram{surfacecolor}{}), outside (\colorpentagram{outsidecolor}{}), or inside (\colorpentagram{insidecolor}{}). Their coordinates, enriched with positional encoding, form the feature \(z\), which serves as input to the SDF network \(\phi\). During training, a clamped L1 loss is applied, while at inference, deviations from learned SDF values are used for anomaly scoring.}
  \label{fig:pipeline}\vspace{-2mm}
\end{figure*}

\section{Related Work}
\label{sec:rel}

\noindent\textbf{3D Anomaly Detection.} Unsupervised 3D anomaly detection\cite{bergmann2023anomaly,tu2024self,zavrtanik2024cheating,zavrtanik2024keep,gu2024rethinking} is critical for applications like quality control and robotics, but 3D data's sparsity, high dimensionality, noise, and arbitrary poses pose challenges.  Many methods use local geometric features and 3D descriptors with KNN \cite{horwitz2023back}, but are noise-sensitive and lack global context.  Reconstruction-based methods (e.g., IMRNet \cite{IMRNet}) struggle with fine details and are computationally expensive. Teacher-student networks \cite{bergmann2023anomaly} require pose consistency.  Methods like AST \cite{rudolph2023asymmetric} miss subtle object anomalies.  Multimodal (e.g., M3DM \cite{wang2023multimodal}) and memory-augmented (e.g., CPMF \cite{cao2023complementary}) methods improve representations but remain local and lack explicit pose handling. EasyNet \cite{chen2023easynet} has a limited receptive field, restricting global shape understanding.  A key limitation is reliance on local features or explicit engineering, causing brittleness to pose variations and hindering generalization.  More complex self-supervised (e.g., R3D-AD~\cite{R3D-AD}) and memory-based (e.g., Reg3D~\cite{Real3d-AD}, Group3AD~\cite{Group3AD}) methods improve generalizability but are computationally expensive.  Directly modeling object geometry offers greater robustness.

\vspace{1mm}
\noindent\textbf{Implicit 3D Representation.} Implicit representations, such as Signed Distance Functions (SDFs) \cite{park2019deepsdf} and occupancy fields \cite{mescheder2019occupancy}, offer compact and continuous descriptions of 3D surfaces, enabling smooth and topologically flexible shape modeling.  They have been applied in shape reconstruction \cite{park2019deepsdf,mescheder2019occupancy}, completion \cite{chibane2020implicit,chou2023diffusionsdf}, and novel view synthesis \cite{long2023neuraludf,zhang2021learning}. Unlike explicit representations (e.g., voxel grids, point clouds, meshes), implicit methods capture fine-grained geometric details without resolution restrictions.  Neural Radiance Fields (NeRF) \cite{mildenhall2021nerf} extend implicit representations to volumetric rendering, but their reliance on dense image supervision and high computational cost limits their applicability to sparse 3D point clouds. Existing 3D anomaly detection methods \cite{Bergmann_2022_mvtec_3dad,IMRNet,Real3d-AD,cao2023complementary,cheng2024towards} often use discrete representations, struggling with fine details and resolution inconsistencies.  We propose an implicit SDF-based approach to address these limitations, leveraging the continuous surface representation of SDFs to preserve global shape consistency and enable accurate anomaly detection and repair.

\vspace{1mm}
\noindent\textbf{3D Anomaly Repair.} 3D anomaly repair is crucial for quality assurance in additive manufacturing (AM), ensuring fabricated parts meet geometric and mechanical integrity standards \cite{1wong2012review}. Traditional open-loop AM processes \cite{4stojadinovic2016towards,5yu2013adaptive} often result in defects due to material deposition inconsistencies, degrading part quality.  Closed-loop AM systems \cite{14liu2019image,15rebaioli2019process}, integrating in-situ 3D scanning with automated defect detection and correction \cite{singh2023toward}, address this issue. Unlike conventional anomaly detection methods \cite{cao2023complementary,Real3d-AD,Group3AD} relying on indirect feature mappings, our PASDF algorithm uses SDF values for direct geometric reasoning, computing anomaly scores.  This enables direct 3D anomaly repair, bridging detection and correction within a unified framework.
\section{Method}
\label{sec:methodl}
\subsection{Problem Statement}
\noindent\textbf{Unsupervised Anomaly Detection.} Let $\mathcal{X} \subset \mathbb{R}^{M \times 3}$ denote the space of 3D point clouds, where each point cloud $Q \in \mathcal{X}$ consists of $M$ points in $\mathbb{R}^3$. Given a category $c$, the goal of \textbf{\textit{unsupervised 3D point cloud anomaly detection}} is to learn a scoring function from a training set composed of normal samples.
Formally, let $\mathcal{S}_c = \{Q_i\}_{i=1}^{N}$ represent the training set for each category $c$, where $Q_i \in \mathcal{X}$ is a normal point cloud. The task is to learn a scoring function $\psi: \mathcal{X} \rightarrow \mathbb{R}^{M}$ that maps an input point cloud $Q \in \mathcal{X}$ to a vector of anomaly scores, $\mathbf{a} = \psi(Q) \in \mathbb{R}^{M}$. Each element $a_j \in \mathbf{a}$ represents the anomaly score for the $j$-th point in the input point cloud $Q$.
The anomaly scores $\mathbf{a} \in \mathbb{R}^{M}$ enable: 1) detection of object-level anomalies through aggregation: $A_{object}(Q) = \text{Aggregate}(\psi(Q))$, and 2) location of pixel-level anomalies using point-level scores $a_j$.

\vspace{0.5mm}
\noindent\textbf{Anomaly Repair.}
The objective of anomaly repair is to learn a repair function $\mathcal{R}: \mathcal{X} \rightarrow \mathcal{X}$ that maps an input point cloud (anomalous) $Q'$ to a repaired point cloud $Q_{repaired}$:
$Q_{repaired} = \mathcal{R}(Q')$.


\subsection{Theoretical Modeling and Intuition}

We employ \textbf{canonical regularization to disentangle pose and shape} for robust 3D anomaly detection. Given a 3D point cloud $Q \in \mathcal{X}$, where $\mathcal{X} \subset \mathbb{R}^{M \times 3}$, we assess the likelihood of a test sample $Q_{test}$ conforming to the normal shape distribution $\mathcal{S}_c$.
To ensure pose invariance, we model this likelihood as: \begin{equation} p(Q \in \mathcal{S}_c) = \int_{SE(3)} p(Q|S, T) p(S \in \mathcal{S}_c) p(T) dT, \end{equation} where $S$ is the intrinsic shape, $T \in SE(3)$ is the pose, and $p(Q|S, T)$ is the likelihood of observing $Q$ given $S$ and $T$.
We approximate this integral by aligning $Q_{test}$ to a canonical pose via: \begin{equation} T^* = \mathop{\text{argmin}}_{T \in SE(3)} \quad d(T(Q_{test}), \mathcal{T}), \end{equation} where $\mathcal{T} \in \mathcal{X}$ is a reference shape and $d(\cdot, \cdot)$ is a distance metric.
This alignment yields $Q'{test} = T^*(Q_{test})$, enabling anomaly detection based on intrinsic shape variations while mitigating pose effects.

\subsection{Pipeline Overview}
We propose an unsupervised 3D point cloud anomaly detection method that learns a consistent, pose-invariant geometric representation via a pose-aligned SDF network. As shown in Fig.~\ref{fig:pipeline}, our pipeline consists of three key stages: 1) \textbf{Pose-wise Alignment Module (PAM)}: This module aligns input point clouds to a canonical coordinate frame, mitigating pose variations. 2) \textbf{Signed Distance Function (SDF) network}: The aligned point clouds are implicitly represented using a SDF network, mapping 3D coordinates (with positional encoding) to signed distances for a smooth and flexible shape representation. 3) \textbf{Anomaly Score Calculation}: At inference, the trained SDF network predicts signed distances for test point clouds, with deviations from expected values serving as point-level anomaly scores, which are aggregated for object-level detection and pixel-level localization.

\begin{algorithm}[t]
  \caption{Pose-wise Alignment Module}
  \label{alg:pose_alignment}
  \begin{algorithmic}[1]
    \Require
      Source point cloud $\mathcal{S}$,
      Target point cloud $\mathcal{T}$,
      Voxel size $v$,
      Loss threshold $\tau$,
      Max iterations $K$
    \Ensure
      Aligned point cloud $\mathcal{S}_\text{aligned}$

    \State Initialize transformation matrix $T \gets \mathbf{I}$
    \State Initialize iteration counter $k \gets 1$

    \While{$k \leq K$}
      \State $\mathcal{S}', \mathcal{T}' \gets \text{Downsample}(\mathcal{S}, \mathcal{T}, v)$
      \State $\mathbf{F}_\mathcal{S}, \mathbf{F}_\mathcal{T} \gets \text{Compute\_FPFH}(\mathcal{S}', \mathcal{T}')$
      \State $T_\text{ransac} \gets \text{RANSAC}(\mathcal{S}', \mathcal{T}',\mathbf{F}_\mathcal{S}, \mathbf{F}_\mathcal{T})$
      \State $T_\text{icp} \gets \text{ICP}(\mathcal{S}, \mathcal{T}, T_\text{ransac})$
      \State $T \gets T_\text{icp} \cdot T_\text{ransac} \cdot T$ 
      \State $\mathcal{S}_\text{aligned} \gets \text{Transform}(\mathcal{S}, T_\text{icp})$
      \State Compute Chamfer Loss $\mathcal{L}$

      \If{$\mathcal{L} < \tau$}
        \State \textbf{break}
      \Else
        \State $\tau \gets \tau + \Delta\tau$
      \EndIf
      \State $k \gets k + 1$
    \EndWhile
    \State \textbf{return} $\mathcal{S}_\text{aligned}$
  \end{algorithmic}
\end{algorithm}


\begin{table*}[th]
  \centering \setlength{\abovecaptionskip}{0.1cm}
  \begin{adjustbox}{max width=\textwidth}
    \begin{tabular}{p{9.315em}|cccccccccccc|c}
    \toprule
    \textbf{(a) O-AUROC~$\uparrow$} & Airplane & Car & Candy & Chicken & Diamond & Duck & Fish & Gemstone & Seahorse & Shell & Starfish & Toffees & \textbf{Mean} \\
    \midrule
   \rowcolor[HTML]{F1F3F4}   BTF(Raw)            & 0.730    & 0.647    & 0.539    & 0.789    & 0.707    & \underline{0.691}    & 0.602    & \textbf{0.686}    & 0.596    & 0.396    & 0.530    & 0.703    & 0.635 \\
    BTF(FPFH)           & 0.520    & 0.560    & 0.630    & 0.432    & 0.545    & \textbf{0.784} & 0.549    & 0.648    & 0.779 & \underline{0.754}    & 0.575    & 0.462    & 0.603 \\
   \rowcolor[HTML]{F1F3F4}   M3DM(PointBERT)     & 0.407    & 0.506    & 0.442    & 0.673    & 0.627    & 0.466    & 0.556    & 0.617    & 0.494    & 0.577    & 0.528    & 0.562    & 0.538 \\
    M3DM(PointMAE)      & 0.434    & 0.541    & 0.552    & 0.683    & 0.602    & 0.433    & 0.540    & 0.644    & 0.495    & 0.694 & 0.551    & 0.450    & 0.552 \\
   \rowcolor[HTML]{F1F3F4}   PatchCore(FPFH)     & \textbf{0.882} & 0.590    & 0.541    & 0.837    & 0.574    & 0.546    & 0.675    & 0.370    & 0.505    & 0.589    & 0.441    & 0.565    & 0.593 \\
    PatchCore(FPFH+Raw) & \underline{0.848} & \underline{0.777} & 0.626    & \textbf{0.853} & 0.784    & 0.628    & 0.837    & 0.359    & 0.767    & 0.663    & 0.471    & 0.570    & 0.682 \\
    \rowcolor[HTML]{F1F3F4}  PatchCore(PointMAE) & 0.726    & 0.498    & 0.663    & 0.827    & 0.783    & 0.489    & 0.630    & 0.374    & 0.539    & 0.501    & 0.519    & 0.585    & 0.594 \\
    RegAD               & 0.716    & 0.697    & 0.685    & \underline{0.852} & 0.900    & 0.584    & 0.915    & 0.417    & 0.762    & 0.583    & 0.506    & \underline{0.827} & 0.704 \\
   \rowcolor[HTML]{F1F3F4}   IMRNet              & 0.762    & 0.711    & 0.755    & 0.780    & \underline{0.905} & 0.517    & 0.880    & \underline{0.674}    & 0.604    & 0.665    & \textbf{0.674} & 0.774    & 0.725 \\
    Group3AD            & 0.744    & 0.728    & \textbf{0.847} & 0.786    & \textbf{0.932} & 0.679    & \underline{0.976} & 0.539    & \underline{0.841}    & 0.585    & 0.562    & 0.796    & \underline{0.751} \\ \midrule
   \rowcolor[HTML]{F1F3F4}   \textbf{PASDF(Ours)}         & 0.628    & \textbf{0.959} & \underline{0.788} & 0.739    & 0.894    & 0.658    & \textbf{0.989} & 0.634    & \textbf{1.000} & \textbf{0.850} & \underline{0.617}    & \textbf{0.866}    & \textbf{0.802} \\
    \midrule\midrule
    \textbf{(b) P-AUROC~$\uparrow$} & Airplane & Car & Candy & Chicken & Diamond & Duck & Fish & Gemstone & Seahorse & Shell & Starfish & Toffees & \textbf{Mean} \\
    \midrule
   \rowcolor[HTML]{F1F3F4}   BTF(Raw)            & 0.564    & 0.647    & 0.735    & 0.609    & 0.563    & 0.601    & 0.514    & 0.597    & 0.520    & 0.489    & 0.392    & 0.623    & 0.571 \\
    BTF(FPFH)           & \underline{0.738} & 0.708    & \textbf{0.864} & 0.735    & \textbf{0.882} & \textbf{0.875}    & 0.709    & \textbf{0.891} & 0.512    & 0.571    & 0.501    & \underline{0.815} & 0.733 \\
  \rowcolor[HTML]{F1F3F4}   M3DM(PointBERT)     & 0.523    & 0.593    & 0.682    & \textbf{0.790} & 0.594    & 0.668    & 0.589    & 0.646    & 0.574    & 0.732    & 0.563    & 0.677    & 0.636 \\
    M3DM(PointMAE)      & 0.530    & 0.607    & 0.683    & 0.735    & 0.618    & 0.678    & 0.600    & 0.654    & 0.561    & 0.748    & 0.555    & 0.679    & 0.637 \\
   \rowcolor[HTML]{F1F3F4}   PatchCore(FPFH)     & 0.471    & 0.643    & 0.637    & 0.618    & 0.760    & 0.430    & 0.464    & \underline{0.830}    & 0.544    & 0.596    & 0.522    & 0.411    & 0.577 \\
    PatchCore(FPFH+Raw) & 0.556    & 0.740    & \underline{0.749}    & 0.558    & 0.854    & 0.658    & 0.781    & 0.539    & 0.808    & 0.753    & 0.613    & 0.549    & 0.680 \\
  \rowcolor[HTML]{F1F3F4}    PatchCore(PointMAE) & 0.579    & 0.610    & 0.635    & 0.683    & 0.776    & 0.439    & 0.714    & 0.514    & 0.660    & 0.725    & \underline{0.641}    & 0.727    & 0.642 \\
    RegAD               & 0.631    & 0.718    & 0.724    & 0.676    & 0.835    & 0.503    & 0.826    & 0.545    & 0.817 & \textbf{0.811} & 0.617    & 0.759    & 0.705 \\
   \rowcolor[HTML]{F1F3F4}   R3D-AD              & 0.730 & 0.647    & 0.539    & \underline{0.789} & 0.707    & 0.691    & 0.602    & 0.686    & 0.596    & 0.396    & 0.530    & 0.703    & 0.635 \\
    Group3AD            & 0.636    & \underline{0.745} & 0.738    & 0.759    & \underline{0.862}    & 0.631    & \underline{0.836 }   & 0.564    & \underline{0.827}    & \underline{0.798}    & 0.625    & 0.803    & \underline{0.735} \\ \midrule
  \rowcolor[HTML]{F1F3F4}    \textbf{PASDF(Ours)}         & \textbf{0.777} & \textbf{0.802} & 0.546    & 0.768    & 0.699    & \underline{0.782} & \textbf{0.837}    & 0.654    & \textbf{0.887} & 0.648    & \textbf{0.703} & \textbf{0.838} & \textbf{0.745} \\
    \bottomrule
    \end{tabular}
  \end{adjustbox}
\caption{Anomaly detection results on Real3D-AD dataset, measured by (a) O-AUROC ($\uparrow$) and (b) P-AUROC ($\uparrow$). The best and second-best results are marked in \textbf{bold} and \underline{underlined}, respectively. Our method achieves the best performance on both metrics.}
\label{Real3D-AD-AUROC}\vspace{-1mm}
\end{table*}

\subsection{Pose-wise Alignment Module}

The PAM estimates the optimal rigid transformation $T^* \in SE(3)$ that aligns the input point cloud $\mathcal{S}$ to the target point cloud $\mathcal{T}$, minimizing a distance metric:
\begin{equation}T^* = \mathop{\text{argmin}}_{T \in SE(3)} d(T(\mathcal{S}), \mathcal{T}).\end{equation}
The PAM integrates global feature-based coarse alignment with local iterative refinement. The process begins by downsampling $\mathcal{S}$ and $\mathcal{T}$ using voxel grids of size $v_{\text{size}}$:
\begin{equation}
\begin{split}
  \mathcal{S}' &= \text{DownSample}(\mathcal{S}, v),  \\
  \mathcal{T}' &= \text{DownSample}(\mathcal{T}, v),
\end{split}
\end{equation}
followed by Fast Point Feature Histogram (FPFH) extraction, where \(\mathbf{F}_\mathcal{S}\) and \(\mathbf{F}_\mathcal{T}\) denote the extracted FPFH feature sets for the downsampled source \(\mathcal{S}'\) and target \(\mathcal{T}'\), respectively.   Global coarse registration is performed using RANSAC\cite{bolles1981ransac} with dual consistency checks:
\begin{equation}
T_{\text{ransac}} = \mathop{\text{argmin}}_{T} \sum_{(\mathbf{p}_i, \mathbf{q}_j) \in \mathcal{C}} \|T(\mathbf{p}_i) - \mathbf{q}_j\|_2,
\end{equation}
where $\mathcal{C}$ denotes FPFH-generated correspondences. 

The Iterative Closest Point (ICP) algorithm\cite{121791} refines the coarse alignment:
\begin{equation}
T_{\text{icp}} = \mathop{\text{argmin}}_{T} \sum_{\mathbf{p}_i \in \mathcal{S}} \min_{\mathbf{q}_j \in \mathcal{T}} \|T(\mathbf{p}_i) - \mathbf{q}_j\|_2^2,
\end{equation}
To avoid local minima, a Chamfer distance-driven feedback mechanism is introduced:
\begin{equation}
\begin{split}
\mathcal{L}_{\text{chamfer}} = 
&\frac{1}{|\mathcal{S}'|} \sum_{\mathbf{p}_i \in \mathcal{S}'} \min_{\mathbf{q}_j \in \mathcal{T}'} \|\mathbf{p}_i - \mathbf{q}_j\|_2^2 \\
&+ \frac{1}{|\mathcal{T}'|} \sum_{\mathbf{q}_j \in \mathcal{T}'} \min_{\mathbf{p}_i \in \mathcal{S}'} \|\mathbf{q}_j - \mathbf{p}_i\|_2^2.
\end{split}
\end{equation}
The cumulative transformation is iteratively updated via:
\begin{equation}
T^{(k)} = T_{\text{icp}}^{(k)} \cdot T_{\text{ransac}}^{(k)} \cdot T^{(k-1)}, \quad k = 1, 2, \dots, K
\end{equation}
where $k$ is the iteration number. The loss threshold $\tau^{(k)}$ is dynamically adapted based on the current Chamfer distance. The iteration terminates if $\mathcal{L}_{\text{chamfer}}^{(k)}  <  \tau^{(k)}$ or reaching $K$ iterations. See Alg.~\ref{alg:pose_alignment} for details.

\subsection{Signed Distance Functions Learning}
After pose alignment, we represent the shape of each object using a Signed Distance Function (SDF) $f_{\theta}: \mathbb{R}^3 \rightarrow \mathbb{R}$, which maps a 3D coordinate $\mathbf{x} \in \mathbb{R}^3$ to its signed distance from the object's surface. The SDF is parameterized by a neural network with parameters $\theta$.

To train the SDF network, we sample query points in the vicinity of the aligned point clouds, categorizing them into surface, external, and internal points. For each query point $\mathbf{x}_i$, we compute a positional encoding $\gamma(\mathbf{x}_i) \in \mathbb{R}^d$ using a sinusoidal encoding function. The SDF network predicts the signed distance value for each query point:
\begin{equation}
\hat{s}_i = f_{\theta}(\gamma(\mathbf{x}_i)).
\end{equation}
The network is trained to minimize a clamped L1 loss:
\begin{equation}
\mathcal{L}_{SDF} = \frac{1}{N_{q}} \sum_{i=1}^{N_{q}} | \text{clamp}(\hat{s}_i, -d_{max}, d_{max}) - s_i |,
\end{equation}
where $s_i$ is the ground truth signed distance, $d_{max}$ is a clamping distance, and $N_q$ is the number of query points. Ground truth signed distances are computed using a nearest neighbor search on the aligned point cloud.




\subsection{Anomaly Score Calculation}
Given a trained SDF network $f_{\theta}$, we define an anomaly score that quantifies the deviation of a test point cloud from the learned normal shape distribution. For a test point cloud $Q_{\text{test}}$, we first align it using the PAM, resulting in $Q'_{\text{test}}$. We then directly input the surface points of $Q'_{\text{test}}$ into the trained SDF network.

The anomaly score for each surface point $\mathbf{x}_j$ is computed as the absolute value of the predicted SDF:
\begin{equation}
A(\mathbf{x}_j) = |f_{\theta}(\mathbf{x}_j)|.
\end{equation}

To obtain the object-level anomaly score, we first select the top-$K$ highest anomaly scores among all points:
\begin{equation}
A_{\text{object}}(Q_{\text{test}}) = \frac{1}{K} \sum_{j \in \mathcal{T}_K} A(\mathbf{x}_j),
\end{equation}
where $\mathcal{T}_K$ denotes the indices of the top-$K$ points with the highest anomaly scores.

\subsection{Anomaly Repair via Learned SDF}
\label{subsec:sdf_repair}

The learned SDF network, $f_{\theta}$, implicitly represents the `normal' shape manifold. This allows for the direct generation of a repaired point cloud, $Q_{repaired}$, given an anomalous input, $Q'$. The repair procedure is as follows:

First, $Q'$ is aligned to the canonical pose via the Pose-wise Alignment Module, resulting in $Q'_{aligned}$. Subsequently, the zero-level set of $f_{\theta}$ is extracted using the Marching Cubes algorithm \cite{lorensen1998marching}. This is performed on a 3D grid that encompasses $Q'_{aligned}$, producing a triangular mesh that approximates the surface defined by $f_{\theta}(\mathbf{x}) = 0$. Finally, a repaired point cloud, $Q_{repaired}$, is obtained by sampling points from this mesh. The process can be formalized as:
\begin{equation}
    Q_{repaired} = \text{SamplePoints}(\text{MarchingCubes}(f_{\theta}, \text{grid}, 0))
\end{equation}
where $\text{MarchingCubes}$ extracts the zero-level isosurface of $f_{\theta}$ evaluated on the specified `grid', and $\text{SamplePoints}$ generates the point cloud from the resulting mesh.

\section{Experiments}

\subsection{Experimental Settings}


\begin{table*}[th]
  \centering   \setlength\tabcolsep{0.8mm}     \setlength{\abovecaptionskip}{0.1cm}
  \begin{adjustbox}{max width=\textwidth}
    \begin{tabular}{l|*{27}{S[table-format=1.3]}}  
    \toprule

    Method & \multicolumn{1}{c}{cap0} & \multicolumn{1}{c}{cap3} & \multicolumn{1}{c}{helmet3} & \multicolumn{1}{c}{cup0} & \multicolumn{1}{c}{bowl4} & \multicolumn{1}{c}{vase3} & \multicolumn{1}{c}{headset1} & \multicolumn{1}{c}{eraser0} & \multicolumn{1}{c}{vase8} & \multicolumn{1}{c}{cap4} & \multicolumn{1}{c}{vase2} & \multicolumn{1}{c}{vase4} & \multicolumn{1}{c}{helmet0} & \multicolumn{1}{c}{bucket1} & \multicolumn{1}{c}{vase7} & \multicolumn{1}{c}{helmet2} & \multicolumn{1}{c}{cap5} & \multicolumn{1}{c}{shelf0} & \multicolumn{1}{c}{bowl5} & \multicolumn{1}{c}{bowl3} & \multicolumn{1}{c}{helmet1} \\
    \midrule
    BTF (Raw)  &    0.668 & 0.527 & 0.526 & 0.403 & 0.664 & 0.717 & 0.515 & 0.525 & 0.424 & 0.468 & 0.410 & 0.425 & 0.553 & 0.321 & 0.448 & 0.602 & 0.373 & 0.164 & 0.417 & 0.385 & 0.349 \\
   \rowcolor[HTML]{F1F3F4}   BTF(FPFH) & 0.618 & 0.522 & 0.444 & 0.586 & 0.609 & 0.699 & 0.490 & 0.719 & 0.668 & 0.520 & 0.546 & 0.510 & 0.571 & 0.633 & 0.518 & 0.542 & 0.586 & 0.609 & 0.699 & 0.490 & 0.719 \\
    M3DM & 0.557 & 0.423 & 0.374 & 0.539 & 0.464 & 0.439 & 0.617 & 0.627 & 0.663 & \textbf{0.777} & 0.737 & 0.476 & 0.526 & 0.501 & 0.657 & 0.623 & 0.639 & 0.564 & 0.409 & 0.617 & 0.427 \\
  \rowcolor[HTML]{F1F3F4}    Patchcore(FPFH)& 0.580 & 0.453 & 0.404 & 0.600 & 0.494 & 0.449 & 0.637 & 0.657 & 0.662 & \underline{0.757} & 0.721 & 0.506 & 0.546 & 0.551 & 0.693 & 0.425 & \underline{0.790} & 0.494 & 0.558 & 0.537 & 0.484 \\
    Patchcore(PointMAE)& 0.589 & 0.476 & 0.424 & 0.610 & 0.501 & 0.460 & 0.627 & 0.677 & 0.663 & 0.727 & 0.741 & 0.516 & 0.556 & 0.561 & 0.650 & 0.447 & 0.538 & 0.523 & 0.593 & 0.579 & 0.552 \\
  \rowcolor[HTML]{F1F3F4}    CPMF& 0.601 & 0.551 & 0.520 & 0.497 & 0.683 & 0.582 & 0.458 & 0.689 & 0.529 & 0.553 & 0.582 & 0.514 & 0.555 & 0.601 & 0.397 & 0.462 & 0.697 & 0.685 & 0.685 & 0.658 & 0.589  \\
    RegAD & 0.693 & 0.725 & 0.367 & 0.510 & 0.663 & 0.650 & 0.610 & 0.343 & 0.620 & 0.643 & 0.605 & 0.500 & 0.600 & 0.752 & 0.462 & 0.614 & 0.467 & 0.688 & 0.593 & 0.348 & 0.381 \\
   \rowcolor[HTML]{F1F3F4}   IMRNet & 0.737 & \textbf{0.775} & 0.573 & 0.643 & 0.676 & 0.700 & \underline{0.676} & 0.548 & 0.630 & 0.652 & 0.614 & 0.524 & 0.597 & \underline{0.771} & 0.635 & \underline{0.641} & 0.652 & 0.603 & \underline{0.710} & 0.599 & 0.600 \\
    R3D-AD & \underline{0.822} & \underline{0.730} & \underline{0.707} & \underline{0.776} & \underline{0.744} & \underline{0.742} & \textbf{0.795} & \underline{0.890} & \underline{0.721} & 0.681 & \underline{0.752} & \underline{0.630} & \underline{0.757} & 0.756 & \underline{0.771} & 0.633 & 0.670 & \underline{0.696} & 0.656 & \underline{0.767} & \underline{0.720} \\ \midrule
  \rowcolor[HTML]{F1F3F4}    \textbf{PASDF(Ours)} & \textbf{0.852} & 0.649 & \textbf{0.846} & \textbf{0.971} & \textbf{0.933} & \textbf{0.806} & \textbf{0.795} & \textbf{0.952} & \textbf{0.924} & 0.646 & \textbf{1.000} & \textbf{0.912} & \textbf{0.812} & \textbf{0.775} & \textbf{1.000} & \textbf{0.765} & \textbf{0.853} & \textbf{0.713} & \textbf{0.912} & \textbf{1.000} & \textbf{0.938} \\
    \midrule \midrule
    Method & \multicolumn{1}{c}{bottle3} & \multicolumn{1}{c}{vase0} & \multicolumn{1}{c}{bottle0} & \multicolumn{1}{c}{tap1} & \multicolumn{1}{c}{bowl0} & \multicolumn{1}{c}{bucket0} & \multicolumn{1}{c}{vase5} & \multicolumn{1}{c}{vase1} & \multicolumn{1}{c}{vase9} & \multicolumn{1}{c}{ashtray0} & \multicolumn{1}{c}{bottle1} & \multicolumn{1}{c}{tap0} & \multicolumn{1}{c}{phone} & \multicolumn{1}{c}{cup1} & \multicolumn{1}{c}{bowl1} & \multicolumn{1}{c}{headset0} & \multicolumn{1}{c}{bag0} & \multicolumn{1}{c}{bowl2} & \multicolumn{1}{c|}{jar0} & \multicolumn{2}{|c}{\textbf{Mean}} \\
    \midrule
    BTF(Raw) & 0.568 & 0.531 & 0.597 & 0.573 & 0.564 & 0.617 & 0.585 & 0.549 & 0.564 & 0.578 & 0.510 & 0.525 & 0.563 & 0.521 & 0.264 & 0.378 & 0.410 & 0.525 & 0.420 & \multicolumn{2}{|c}{0.493} \\
   \rowcolor[HTML]{F1F3F4}   BTF(FPFH) & 0.322 & 0.342 & 0.344 & 0.546 & 0.509 & 0.401 & 0.409 & 0.219 & 0.268 & 0.420 & 0.546 & 0.560 & 0.671 & 0.610 & 0.668 & 0.520 & 0.546 & 0.510 & 0.424 & \multicolumn{2}{|c}{0.528} \\
    M3DM & 0.541 & 0.423 & 0.574 & 0.739 & 0.634 & 0.309 & 0.317 & 0.427 & 0.663 & 0.577 & 0.637 & \underline{0.754} & 0.357 & 0.556 & 0.663 & 0.577 & 0.537 & 0.684 & 0.441 & \multicolumn{2}{|c}{0.552} \\
    \rowcolor[HTML]{F1F3F4}  Patchcore(FPFH) & 0.572 & 0.455 & 0.604 & 0.766 & 0.504 & 0.469 & 0.417 & 0.423 & 0.660 & 0.587 & 0.667 & 0.753 & 0.388 & 0.586 & 0.639 & 0.583 & 0.571 & 0.615 & 0.472 & \multicolumn{2}{|c}{0.568} \\
    Patchcore(PointMAE) & 0.650 & 0.447 & 0.513 & 0.538 & 0.523 & 0.593 & 0.579 & 0.552 & 0.629 & 0.591 & 0.601 & 0.458 & 0.488 & 0.556 & 0.629 & 0.591 & 0.601 & 0.458 & 0.483 & \multicolumn{2}{|c}{0.562} \\
   \rowcolor[HTML]{F1F3F4}   CPMF & 0.405 & 0.451 & 0.520 & 0.697 & 0.783 & 0.482 & 0.618 & 0.345 & 0.609 & 0.353 & 0.482 & 0.359 & 0.509 & 0.499 & 0.639 & 0.643 & 0.643 & 0.625 & 0.610 & \multicolumn{2}{|c}{0.559}\\
    RegAD & 0.525 & 0.533 & 0.486 & 0.641 & 0.671 & 0.610 & 0.520 & 0.702 & 0.594 & 0.597 & 0.695 & 0.676 & 0.414 & 0.538 & 0.525 & 0.537 & 0.706 & 0.490 & 0.592 & \multicolumn{2}{|c}{0.572} \\
   \rowcolor[HTML]{F1F3F4}   IMRNet & 0.640 & 0.533 & 0.552 & 0.696 & 0.681 & 0.580 & 0.676 & \underline{0.757} & 0.594 & 0.671 & 0.700 & 0.676 & 0.755 & \underline{0.757} & 0.702 & 0.720 & 0.660 & 0.685 & 0.780 & \multicolumn{2}{|c}{0.661} \\
    R3D-AD & \underline{0.781} & \underline{0.788} & \underline{0.733} & \textbf{0.900} & \underline{0.819} & \underline{0.683} & \underline{0.757} & 0.729 & \underline{0.718} & \underline{0.833} & \underline{0.737} & 0.736 & \underline{0.762} & \underline{0.757} & \underline{0.778} & \underline{0.738} & \underline{0.720} & \underline{0.741} & \underline{0.838} & \multicolumn{2}{|c}{\underline{0.749}} \\ \midrule
  \rowcolor[HTML]{F1F3F4}    \textbf{PASDF(Ours)} & \textbf{1.000} & \textbf{1.000} & \textbf{1.000} & \underline{0.793} & \textbf{1.000} & \textbf{0.968} & \textbf{1.000} & \textbf{0.929} & \textbf{0.836} & \textbf{1.000} & \textbf{1.000} & \textbf{0.882} & \textbf{1.000} & \textbf{0.857} & \textbf{0.948} & \textbf{1.000} & \textbf{0.995} & \textbf{1.000} & \textbf{1.000} & \multicolumn{2}{|c}{\textbf{0.900}} \\
    \bottomrule
    \end{tabular}%
  \end{adjustbox}
  \caption{O-AUROC score~($\uparrow$) on Anomaly-ShapeNet dataset. The best and second-best results are marked in \textbf{bold} and \underline{underlined}.
  }
  \label{Anomaly-ShapeNet-O-AUROC}\vspace{-2mm}
\end{table*}

\begin{table*}[th]
  \centering     \setlength{\abovecaptionskip}{0.1cm}
  \setlength\tabcolsep{0.8mm}
  \begin{adjustbox}{max width=\textwidth}
    \begin{tabular}{l|*{27}{S[table-format=1.3]}}
    \toprule
Method & \multicolumn{1}{c}{cap0} & \multicolumn{1}{c}{cap3} & \multicolumn{1}{c}{helmet3} & \multicolumn{1}{c}{cup0} & \multicolumn{1}{c}{bowl4} & \multicolumn{1}{c}{vase3} & \multicolumn{1}{c}{headset1} & \multicolumn{1}{c}{eraser0} & \multicolumn{1}{c}{vase8} & \multicolumn{1}{c}{cap4} & \multicolumn{1}{c}{vase2} & \multicolumn{1}{c}{vase4} & \multicolumn{1}{c}{helmet0} & \multicolumn{1}{c}{bucket1} & \multicolumn{1}{c}{vase7} & \multicolumn{1}{c}{helmet2} & \multicolumn{1}{c}{cap5} & \multicolumn{1}{c}{shelf0} & \multicolumn{1}{c}{bowl5} & \multicolumn{1}{c}{bowl3} & \multicolumn{1}{c}{helmet1} \\
    \midrule
  \rowcolor[HTML]{F1F3F4}    BTF(Raw) & 0.524 & 0.687 & 0.700 & 0.632 & 0.563 & 0.602 & 0.475 & 0.637 & 0.550 & 0.469 & 0.403 & 0.613 & 0.504 & 0.686 & 0.578 & 0.605 & 0.373 & 0.464 & 0.517 & 0.685 & 0.449 \\
    BTF(FPFH) & \underline{0.730} & 0.658 & 0.724 & \underline{0.790} & 0.679 & \underline{0.699} & 0.591 & 0.719 & 0.662 & 0.524 & 0.646 & 0.710 & 0.575 & 0.633 & 0.540 & 0.643 & 0.586 & 0.619 & 0.699 & \underline{0.690} & \textbf{0.749} \\
  \rowcolor[HTML]{F1F3F4}    M3DM & 0.531 & 0.605 & 0.655 & 0.715 & 0.624 & 0.658 & 0.585 & 0.710 & 0.551 & 0.718 & 0.737 & 0.655 & 0.599 & 0.699 & 0.517 & 0.623 & 0.655 & 0.554 & 0.489 & 0.657 & 0.427 \\
    Patchcore(FPFH) & 0.472 & 0.653 & \underline{0.737} & 0.655 & 0.720 & 0.430 & 0.464 & \underline{0.810} & 0.575 & 0.595 & 0.721 & 0.505 & 0.548 & 0.571 & 0.693 & 0.455 & \underline{0.795} & 0.613 & 0.358 & 0.327 & 0.489 \\
  \rowcolor[HTML]{F1F3F4}    Patchcore(PointMAE) & 0.544 & 0.488 & 0.615 & 0.510 & 0.501 & 0.465 & 0.423 & 0.378 & 0.364 & 0.725 & \underline{0.742} & 0.523 & 0.580 & 0.754 & 0.651 & 0.651 & 0.545 & 0.543 & 0.562 & 0.581 & 0.562 \\
    CPMF & 0.601 & 0.551 & 0.520 & 0.497 & 0.683 & 0.582 & 0.458 & 0.689 & 0.529 & 0.553 & 0.582 & 0.514 & 0.555 & 0.601 & 0.504 & 0.515 & 0.551 & \underline{0.783} & 0.684 & 0.641 & 0.542 \\
   \rowcolor[HTML]{F1F3F4}   RegAD & 0.632 & \underline{0.718} & 0.620 & 0.685 & \underline{0.800} & 0.511 & \underline{0.626} & 0.755 & \underline{0.811} & \underline{0.815} & 0.405 & \underline{0.755} & \underline{0.600} & 0.725 & \underline{0.881} & \textbf{0.825} & 0.467 & 0.688 & 0.691 & 0.654 & 0.624 \\
    IMRNet & 0.715 & 0.706 & 0.663 & 0.643 & 0.576 & 0.401 & 0.476 & 0.548 & 0.635 & 0.753 & 0.614 & 0.524 & 0.598 & \underline{0.774} & 0.593 & 0.644 & 0.742 & 0.605 & \underline{0.715} & 0.599 & 0.604 \\ \midrule
  \rowcolor[HTML]{F1F3F4}    \textbf{PASDF(Ours)} & \textbf{0.948} & \textbf{0.861} & \textbf{0.958} & \textbf{0.948} & \textbf{0.865} & \textbf{0.868} & \textbf{0.891} & \textbf{0.945} & \textbf{0.909} & \textbf{0.894} & \textbf{0.956} & \textbf{0.899} & \textbf{0.816} & \textbf{0.824} & \textbf{0.959} & \underline{0.809} & \textbf{0.920} & \textbf{0.865} & \textbf{0.909} & \textbf{0.939} & \underline{0.646} \\
    \midrule \midrule
Method & \multicolumn{1}{c}{bottle3} & \multicolumn{1}{c}{vase0} & \multicolumn{1}{c}{bottle0} & \multicolumn{1}{c}{tap1} & \multicolumn{1}{c}{bowl0} & \multicolumn{1}{c}{bucket0} & \multicolumn{1}{c}{vase5} & \multicolumn{1}{c}{vase1} & \multicolumn{1}{c}{vase9} & \multicolumn{1}{c}{ashtray0} & \multicolumn{1}{c}{bottle1} & \multicolumn{1}{c}{tap0} & \multicolumn{1}{c}{phone} & \multicolumn{1}{c}{cup1} & \multicolumn{1}{c}{bowl1} & \multicolumn{1}{c}{headset0} & \multicolumn{1}{c}{bag0} & \multicolumn{1}{c}{bowl2} & \multicolumn{1}{c|}{jar0} & \multicolumn{2}{|c}{\textbf{Mean}} \\ 
    \midrule
   \rowcolor[HTML]{F1F3F4}   BTF(Raw) & \underline{0.720} & 0.618 & 0.551 & 0.564 & 0.524 & 0.617 & 0.585 & 0.549 & 0.564 & 0.512 & 0.491 & 0.527 & 0.583 & 0.561 & 0.464 & 0.578 & 0.430 & 0.426 & 0.423 & \multicolumn{2}{|c}{0.550} \\
    BTF(FPFH) & 0.622 & 0.642 & 0.641 & 0.596 & 0.710 & 0.401 & 0.429 & 0.619 & 0.568 & 0.624 & 0.549 & 0.568 & 0.675 & 0.619 & \underline{0.768} & 0.620 & \underline{0.746} & 0.518 & 0.427 & \multicolumn{2}{|c}{0.628} \\
  \rowcolor[HTML]{F1F3F4}    M3DM & 0.532 & 0.608 & 0.663 & 0.712 & 0.658 & \underline{0.698} & 0.642 & 0.602 & 0.663 & 0.577 & 0.637 & 0.654 & 0.358 & 0.556 & 0.663 & 0.581 & 0.637 & \underline{0.694} & 0.541 & \multicolumn{2}{|c}{0.616} \\
    Patchcore(FPFH) & 0.512 & 0.655 & 0.654 & \underline{0.768} & 0.524 & 0.459 & 0.447 & 0.453 & 0.663 & 0.597 & 0.687 & 0.733 & 0.488 & 0.596 & 0.531 & 0.583 & 0.574 & 0.625 & 0.478 & \multicolumn{2}{|c}{0.580} \\
   \rowcolor[HTML]{F1F3F4}   Patchcore(PointMAE) & 0.653 & \underline{0.677} & 0.553 & 0.541 & 0.527 & 0.586 & 0.572 & 0.551 & 0.423 & 0.495 & 0.606 & \underline{0.858} & \underline{0.886} & \underline{0.856} & 0.524 & 0.575 & 0.674 & 0.515 & 0.487 & \multicolumn{2}{|c}{0.577} \\
    CPMF & 0.435 & 0.458 & 0.521 & 0.657 & 0.745 & 0.486 & 0.651 & 0.486 & 0.545 & 0.615 & 0.571 & 0.458 & 0.545 & 0.509 & 0.488 & \underline{0.699} & 0.655 & 0.635 & 0.611 & \multicolumn{2}{|c}{0.573} \\
   \rowcolor[HTML]{F1F3F4}   RegAD & 0.525 & 0.548 & \underline{0.888} & 0.741 & 0.775 & 0.619 & 0.624 & 0.602 & \underline{0.694} & \underline{0.698} & 0.696 & 0.589 & 0.599 & 0.698 & 0.645 & 0.580 & 0.715 & 0.593 & 0.599 & \multicolumn{2}{|c}{\underline{0.668}} \\
    IMRNet & 0.641 & 0.535 & 0.556 & 0.699 & \underline{0.781} & 0.585 & \underline{0.682} & \underline{0.685} & 0.691 & 0.671 & \underline{0.702} & 0.681 & 0.742 & 0.688 & 0.705 & 0.615 & 0.668 & 0.684 & \underline{0.765} & \multicolumn{2}{|c}{0.650} \\ \midrule
  \rowcolor[HTML]{F1F3F4}    \textbf{PASDF(Ours)} & \textbf{0.948} & \textbf{0.944} & \textbf{0.951} & \textbf{0.902} & \textbf{0.963} & \textbf{0.875} & \textbf{0.915} & \textbf{0.797} & \textbf{0.863} & \textbf{0.919} & \textbf{0.926} & \textbf{0.884} & \textbf{0.951} & \textbf{0.884} & \textbf{0.900} & \textbf{0.863} & \textbf{0.958} & \textbf{0.816} & \textbf{0.959} & \multicolumn{2}{|c}{\textbf{0.897}} \\
    \bottomrule
    \end{tabular}%
  \end{adjustbox}
  \caption{P-AUROC score~($\uparrow$) on Anomaly-ShapeNet dataset. The best and second-best results are marked in \textbf{bold} and \underline{underlined}.
  }
  \label{Anomaly-ShapeNet-P-AUROC}\vspace{-2mm}
\end{table*}

\noindent\textbf{Datasets.}  
We evaluate our method on two benchmark datasets: Real3D-AD~\cite{Real3d-AD} and Anomaly-ShapeNet~\cite{IMRNet}.  
\begin{itemize}
    \item \textbf{Real3D-AD} is a high-resolution real-world dataset with 12 categories, each containing 4 normal training samples and 100 test samples. The test set includes both normal and anomalous objects.  
    \item \textbf{Anomaly-ShapeNet} is a synthetic dataset derived from ShapeNet, comprising 40 categories and over 1,600 samples. Both datasets use complete point clouds for training, but Real3D-AD test samples are single-view, making anomaly detection more challenging.  
\end{itemize}

\noindent\textbf{Baseline Methods.}  
We compare our approach against state-of-the-art methods, including BTF\cite{horwitz2023back}, M3DM\cite{M3DM}, PatchCore\cite{roth2022towards}, CPMF\cite{cao2023complementary}, Reg3D-AD\cite{Real3d-AD}, IMRNet\cite{IMRNet}, R3D-AD\cite{R3D-AD}, and Group3AD\cite{Group3AD}. PatchCore, originally designed for 2D anomaly detection, is extended to 3D by modifying its feature extractor. BTF and PatchCore have different  feature extractor variants. Results are obtained from publicly available code or papers.  

\noindent\textbf{Implementation Details.}  
The PAM module is configured with a loss threshold $ \tau = 0.016 $, an increment $ \Delta\tau = 0.001 $, and a maximum iteration count $ K = 10 $. The SDF network consists of an 8-layer MLP with fully connected layers and weight normalization. Intermediate layers use ReLU activation and 0.2 dropout, except for the final layer, with a skip connection at the fourth layer. The SDF network is trained using clamped L1 loss with a maximum distance threshold of $d_{\max} = 0.1$.  
For SDF learning, we sample 23,000 query points: 10,000 from the surface, 10,000 within the object’s bounding box, and 3,000 within the unit volume. Training runs for 2,000 epochs with $lr=1 \times 10^{-5}$. In the anomaly score computation, $\mathcal{T}_K$ is set to 1000. All experiments are conducted on one NVIDIA V100 GPU.

\subsection{Evaluation Metrics}

\noindent\textbf{Anomaly Detection.}
Following \cite{bergmann2023anomaly}, we use Area Under the Receiver Operating Characteristic Curve (AUROC) to assess anomaly detection. Pixel-AUROC (P-AUROC) evaluates point-level localization, while Object-AUROC (O-AUROC) measures object-level detection. Higher AUROC values indicate better anomaly detection quality.

\noindent\textbf{Anomaly Repair.} 
To assess 3D anomaly repair using the learned SDF, we employ Chamfer Distance (CD) and Earth Mover’s Distance (EMD). CD quantifies the average squared Euclidean distance between reconstructed and ground truth geometry, ensuring geometric accuracy. EMD captures structural differences by measuring the minimal transformation cost between distributions. 

\begin{figure}[t]
  \centering
 \setlength{\abovecaptionskip}{0.0cm}
  \includegraphics[width=0.45\textwidth]{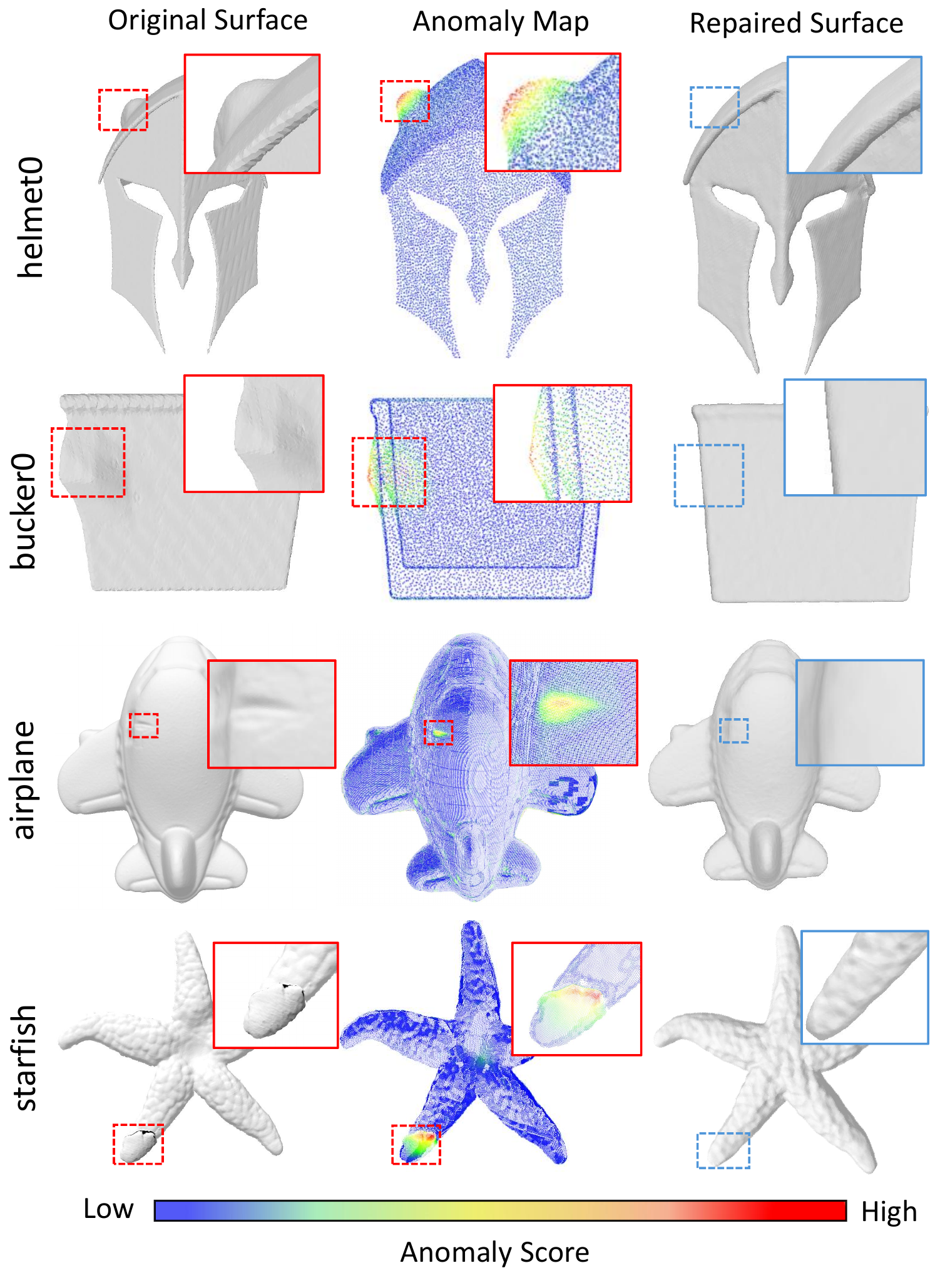}
\caption{\textcolor{black}{Qualitative anomaly localization results on Anomaly-ShapeNet and Real3D-AD (rows 1-2 and 3-4 respectively).}}
  \label{fig:Vis}\vspace{-1mm}
\end{figure}%

\subsection{Main Results}
\textbf{Comparisons on Real3D-AD.} Table \ref{Real3D-AD-AUROC} reports anomaly detection results on Real3D-AD, evaluated by O-AUROC (object-level) and P-AUROC (point-level). Our method, PASDF, achieves the highest mean scores on both metrics, demonstrating superior detection and localization. \textit{\textbf{The proposed PASDF leads with a 0.802 O-AUROC, ranking first in five categories, and a 0.745 P-AUROC, excelling in six.}} It performs particularly well in categories with complex surface textures (e.g., Seahorse, Shell, Starfish, Toffees), where patch-wise methods struggle with fine-grained details. By effectively capturing intricate geometric structures (Fig. \ref{fig:reconstruction}), PASDF enables more precise anomaly detection and localization.

\vspace{1mm}
\noindent\textbf{Comparisons on Anomaly-ShapeNet.} Table \ref{Anomaly-ShapeNet-O-AUROC} reports O-AUROC results on Anomaly-ShapeNet, where PASDF achieves\textit{ \textbf{the highest mean score of 0.900}, \textbf{ranking first in \textbf{37 of 40} categories}}. This highlights its robustness across diverse object types. 
Other methods show performance inconsistencies, excelling in some categories but varying in others. For instance, IMRNet scores 0.774 on \textit{bucket0} but only 0.401 on \textit{vase3}, revealing its instability. In contrast, PASDF consistently delivers strong results across all categories, ensuring greater reliability in anomaly localization. Due to space constraints, detailed object-level results are in the  Appendix of the supplementary material.
Across both datasets, PASDF’s ability to learn consistent geometric representations enables fine-grained texture capture while maintaining strong robustness across object categories.

\subsection{Qualitative Results}
Fig.~\ref{fig:Vis} presents qualitative anomaly localization results on Real3D-AD and Anomaly-ShapeNet. Detected anomalies are highlighted with red dashed circles. Our method effectively identifies structural defects, including deformations and geometric irregularities, across both real-world and synthetic datasets. These qualitative results demonstrate the robustness and generalizability of our anomaly detection approach.

\begin{table}[t]
  \centering  \setlength\tabcolsep{3mm}     \setlength{\abovecaptionskip}{0.1cm}
  \resizebox{0.48\textwidth}{!}{ 
  \begin{tabular}{l|cc|cc}
    \toprule
    \multirow{2}{*}{\textbf{Method}} & \multicolumn{2}{c|}{\textbf{Real3D-AD}} & \multicolumn{2}{c}{\textbf{Anomaly-ShapeNet}} \\
    \cmidrule(lr){2-3} \cmidrule(lr){4-5}
    & \textbf{CD~$\downarrow$} & \textbf{EMD~$\downarrow$} & \textbf{CD~$\downarrow$} & \textbf{EMD~$\downarrow$} \\
    \midrule

    w/o $PE$ & 0.0255 & 0.0133 & 0.0575 & 0.0276  \\ \midrule
  \rowcolor[HTML]{F1F3F4}    with $PE$ & \textbf{0.0203} & \textbf{0.0110} & \textbf{0.0445} & \textbf{0.0228} \\
    \midrule
  \end{tabular}
  }
  \caption{Anomaly repair results. $PE$: positional encoding.  } 
  \label{ablation_cd_emd}\vspace{-1mm}
\end{table}
\begin{table}[t]
  \centering \setlength\tabcolsep{1mm}     \setlength{\abovecaptionskip}{0.1cm}
  \resizebox{\linewidth}{!}{
  \begin{tabular}{l|c|cc}
    \toprule
{\textbf{Method}} & {\textbf{PAM}}  &\textbf{O-AUROC~$\uparrow$} & \textbf{P-AUROC~$\uparrow$}\\
    
    \midrule
    \multirow{2}{*}{BTF(FPFH)}          &  &   0.528 & 0.628 \\
                                         & \checkmark &  0.579 & 0.683 \\ \hline
    \multirow{2}{*}{Patchcore(FPFH)}    &  &  0.568 & 0.580 \\
                                         & \checkmark &  0.814 & 0.867 \\ \hline
    \multirow{2}{*}{Patchcore(PointMAE)}&  &   0.562 & 0.577 \\
                                         & \checkmark  &  0.626 & 0.681 \\ \midrule
     \rowcolor[HTML]{F1F3F4}  {PASDF} (\textbf{Ours})                      &\checkmark & \textbf{0.900} & \textbf{0.897} \\
    \bottomrule
  \end{tabular}
  }
  \caption{ Generalization of PAM for anomaly detection enhancement across diverse baseline models on Anomaly-ShapeNet. }
  \label{ablation-PAM}\vspace{-1mm}
\end{table}
\begin{table}[t]
  \centering \setlength\tabcolsep{4mm}         \setlength{\abovecaptionskip}{0.1cm} \setlength{\abovecaptionskip}{0.1cm}
  \resizebox{1\linewidth}{!}{
  \begin{tabular}{l|cc}
    \toprule
     \textbf{Method} &\textbf{O-AUROC~$\uparrow$} & \textbf{P-AUROC~$\uparrow$}\\
    \midrule

    w/o RANSAC & 0.711 & 0.739 \\
w/o ICP  & 0.727 & 0.836 \\
    w/o $IO$  & 0.871 & 0.884 \\
    w/o $PE$ & 0.887 & 0.783  \\ \midrule
   \rowcolor[HTML]{F1F3F4}   {PASDF} (\textbf{Full})    & \textbf{0.900} & \textbf{0.897} \\
    \midrule
  \end{tabular}
  }
  \caption{Ablation study on Anomaly-ShapeNet. $IO$ represent iterative optimization. $PE$ stands for positional encoding.}
  \label{ablation-submodules}\vspace{-1mm}
\end{table}

\begin{figure*}[t!]
  \centering
  \setlength{\abovecaptionskip}{0.1cm}
  \includegraphics[width=0.89\textwidth]{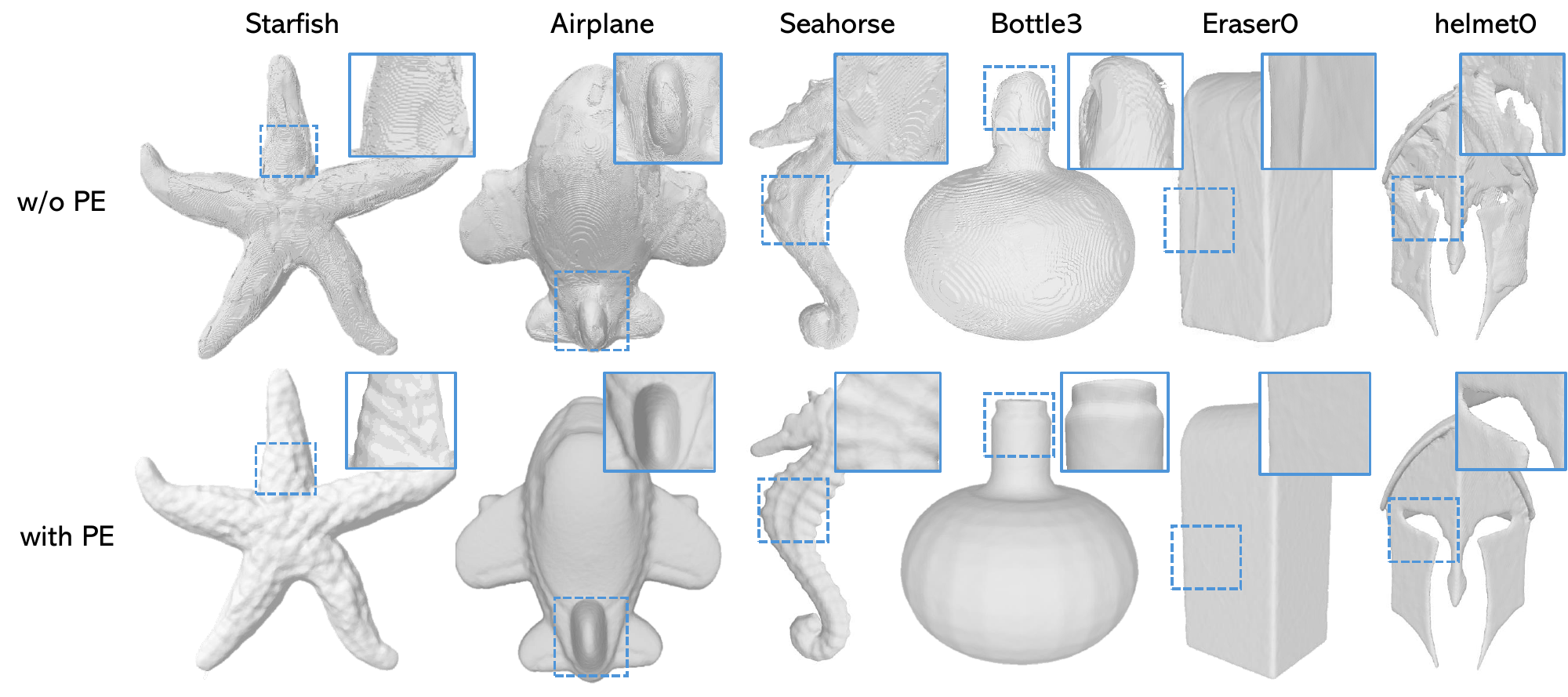}
  \caption{\textcolor{black}{{Qualitative evaluation of PASDF for reconstruction. The first three columns are from the Real3D AD dataset, and the last three columns are from Anomaly-ShapeNet. w/o $PE$ denotes PASDF without positional encoding}
}}
  \label{fig:reconstruction}\vspace{-1mm}
\end{figure*}
\subsection{Ablation Study}  

\noindent\textbf{Effect of Pose-Wise Alignment on Anomaly Detection.}  
Table~\ref{ablation-PAM} shows that the proposed pose alignment module (PAM) consistently enhances anomaly detection across diverse baseline models.  
For BTF(FPFH), PAM improves O-AUROC and P-AUROC by 9.7\% and 8.8\%, respectively. Notably, PatchCore(FPFH) achieves substantial gains of 24.6\% (O-AUROC) and 28.7\% (P-AUROC) with PAM, indicating that pose standardization significantly benefits shallow feature representations by reducing geometric ambiguity. Even for deep learning-based PointMAE features, PAM yields a 6.4\%/10.4\% improvement, demonstrating its adaptability to both handcrafted and learned features.  
These consistent improvements confirm that canonical pose alignment provides a unified geometric reference, allowing models to focus on intrinsic structural characteristics rather than pose variations.

\vspace{1mm}
\noindent\textbf{Effect of Positional Encoding.}  
Ablation studies confirm the crucial role of positional encoding in 3D surface reconstruction and anomaly detection. Using Marching Cubes~\cite{lorensen1998marching}, we reconstruct normal object surfaces as repair templates from SDF values and compare them to ground-truth point clouds in canonical poses. Quantitative results (Table~\ref{ablation_cd_emd}) show that positional encoding improves reconstruction accuracy, reducing Chamfer Distance (CD) and Earth Mover’s Distance (EMD) for better geometric alignment.  
Qualitatively, as shown in Figure~\ref{fig:reconstruction}, models without positional encoding (w/o PE) generate artifacts such as terraced surfaces (e.g., starfish) and lose fine details, whereas those with positional encoding produce smoother, more consistent shapes.  
For anomaly detection (Table~\ref{ablation-PAM}), removing positional encoding reduces P-AUROC by 11.4\% (89.7\% → 78.3\%) and slightly lowers O-AUROC, demonstrating its importance in capturing spatial awareness for detecting fine-grained anomalies.


\vspace{1mm}
\noindent\textbf{Component-Level Analysis.}  
To assess the contributions of individual PAM components, we perform an ablation study (Table~\ref{ablation-submodules}). Removing RANSAC (w/o RANSAC) causes a sharp O-AUROC drop from 0.900 to 0.711, underscoring its importance for robust initial alignment. After RANSAC, ICP further improves P-AUROC from 0.836 to 0.897, highlighting its role in fine-grained pose refinement. The iterative optimization(w/o IO) is also essential, as its removal reduces both O-AUROC and P-AUROC, confirming the necessity of multiple refinement steps for achieving an accurate canonical pose.  
The strong performance of PASDF reaffirms that pose-wise alignment is a vital preprocessing step for robust 3D anomaly representations.




\section{Conclusion}
We introduce PASDF, a novel framework for unsupervised 3D anomaly detection using pose-aware Signed Distance Functions. By disentangling pose and shape through a Pose-wise Alignment Module, PASDF achieves state-of-the-art performance, demonstrating robustness to pose variations and sensitivity to subtle geometric anomalies. Our results highlight the superiority of learning a continuous, pose-invariant 3D representation over discretized approaches that struggle with random pose effects. Additionally, the high-quality 3D template learned via our SDF enables effective anomaly repair.

\vspace{1mm}
\noindent\textbf{Limitations and future work.} PASDF achieves promising results, but limitations suggest future research. The Pose-wise Alignment Module can be computationally expensive under challenging initial pose; future work could explore efficient learning-based or hierarchical registration. PASDF currently assumes a single normal object class; extending it to multi-class detection would broaden applicability. Performance also depends on input point cloud quality, motivating noise and outlier robustness research. Beyond these improvements, integrating PASDF with contextual information could enhance detection in real-world complex environments. 

\clearpage

{
    \small
    \bibliographystyle{ieeenat_fullname}
    \bibliography{ICCV2025-Author-Kit-Feb/main}
}
\clearpage
\setcounter{page}{1}
\maketitlesupplementary

\setcounter{table}{0}
\setcounter{figure}{0}
\setcounter{section}{0}
\renewcommand{\thetable}{\Alph{table}}
\renewcommand{\thefigure}{\Alph{figure}}
\renewcommand{\thesection}{\Alph{section}}

\section{Data Preparation Details} \label{sec:app-data-prep}

Our pipeline processes raw 3D meshes into signed distance function (SDF) training samples through geometric normalization and multi-scale sampling. 
During training, each input mesh undergoes coordinate normalization to align all axes within the [0,1]³ domain. We validate geometric integrity using a hybrid watertightness check: First, we detect non-manifold edges through topological analysis of the face adjacency graph. Meshes failing this check are processed through Poisson surface reconstruction to generate watertight counterparts. 

Query points are sampled through a multi-scale strategy:  \textbf{(1) global volume sampling} across the normalized [0,1]³ domain generates global context points,  \textbf{(2) adaptive bounding box sampling} operates within a 1.3× expanded object-aligned bounding box to concentrate queries in surface-proximal regions, and \textbf{(3) surface-constrained sampling} employs barycentric coordinate interpolation on mesh faces, prioritizing high-curvature regions through triangle area-weighted selection to densely encode the SDF zero-crossing manifold. 
The signed distance value for a query point \(\bm{x}\) is:
\begin{equation}
\text{SDF}(\bm{x}) = \|\bm{x} - \bm{p}_s\|_2 \cdot \text{sgn}\!\left( \bm{n}_s^\top (\bm{x} - \bm{p}_s) \right)
\label{eq:sdf}
\end{equation}
where \(\bm{p}_s\) is the nearest surface point to \(\bm{x}\) (retrieved via KD-tree acceleration), \(\|\bm{x} - \bm{p}_s\|_2\) is the absolute distance from \(\bm{x}\) to the surface,  \(\bm{n}_s\) denotes the outward-oriented unit normal at \(\bm{p}_s\), and \(\text{sgn}(\cdot)\) enforces the geometric sign convention:

\begin{equation}
\text{sgn}(v) = \begin{cases} 
-1 & \text{if } v < 0 \quad (\textit{inside}) \\
+1 & \text{otherwise} \quad (\textit{outside}).
\end{cases}
\end{equation}

The sign assignment follows a consistent logic: points located inside the shape reside in the negative half-space defined by the surface normal, satisfying \(\bm{n}_s^\top (\bm{x} - \bm{p}_s) < 0\), while those in the positive half-space are classified as outside. This approach ensures alignment with the assumption that the mesh is watertight. Furthermore, it is well-suited for point cloud representations that include internal structures, as the explicit separation of distance and sign enables robust handling of enclosed cavities and nested surfaces, ensuring accurate SDF computation even in complex, multi-layered geometries.

\section{Evaluation Metrics for Anomaly Repair} \label{sec:metrics}

To quantitatively evaluate 3D anomaly repair, we use Chamfer Distance and Earth Mover’s Distance, which assess geometric accuracy and structural consistency.

\vspace{1mm}
\noindent\textbf{Chamfer Distance (CD).}
Chamfer Distance measures the discrepancy between two point sets by computing the sum of squared distances from each point to its nearest neighbor in the other set. Given a reconstructed shape \( S_1 \) and the ground truth \( S_2 \), it is defined as:

\begin{equation}
    d_{\text{CD}}(S_1, S_2) = \sum_{x \in S_1} \min_{y \in S_2} \|x - y\|_2^2 + \sum_{y \in S_2} \min_{x \in S_1} \|x - y\|_2^2.
\end{equation}

Here, \( x \in S_1 \) and \( y \in S_2 \) are points in the reconstructed and ground truth point sets, respectively, and \( \| \cdot \|_2 \) denotes the Euclidean distance. CD efficiently captures point-wise accuracy, making it suitable for evaluating local geometric deviations. However, it does not impose a structured mapping between the distributions of \( S_1 \) and \( S_2 \), potentially leading to misalignment in cases where shapes exhibit global structural shifts.

\vspace{1mm}
\noindent\textbf{Earth Mover’s Distance (EMD).}
Earth Mover’s Distance, also known as the Wasserstein distance, measures the minimum transport cost required to transform one point set into another. Unlike CD, which only considers nearest neighbors, EMD enforces a one-to-one correspondence by finding an optimal bijection \( \phi: S_1 \to S_2 \):

\begin{equation}
    d_{\text{EMD}}(S_1, S_2) = \min_{\phi: S_1 \rightarrow S_2} \sum_{x \in S_1} \| x - \phi(x) \|_2.
\end{equation}

In this formulation, \( \phi(x) \) represents the optimal match for each \( x \) in the ground truth set \( S_2 \), ensuring that mass transport is minimized. EMD provides a more global assessment of structural consistency, penalizing uneven distributions that Chamfer Distance might overlook. Due to its higher computational complexity, it is typically computed on a subset of points.

\section{More Qualitative Results}
\begin{figure*}[t!]
  \centering
  \setlength{\abovecaptionskip}{0.1cm}
  \includegraphics[width=1\textwidth]{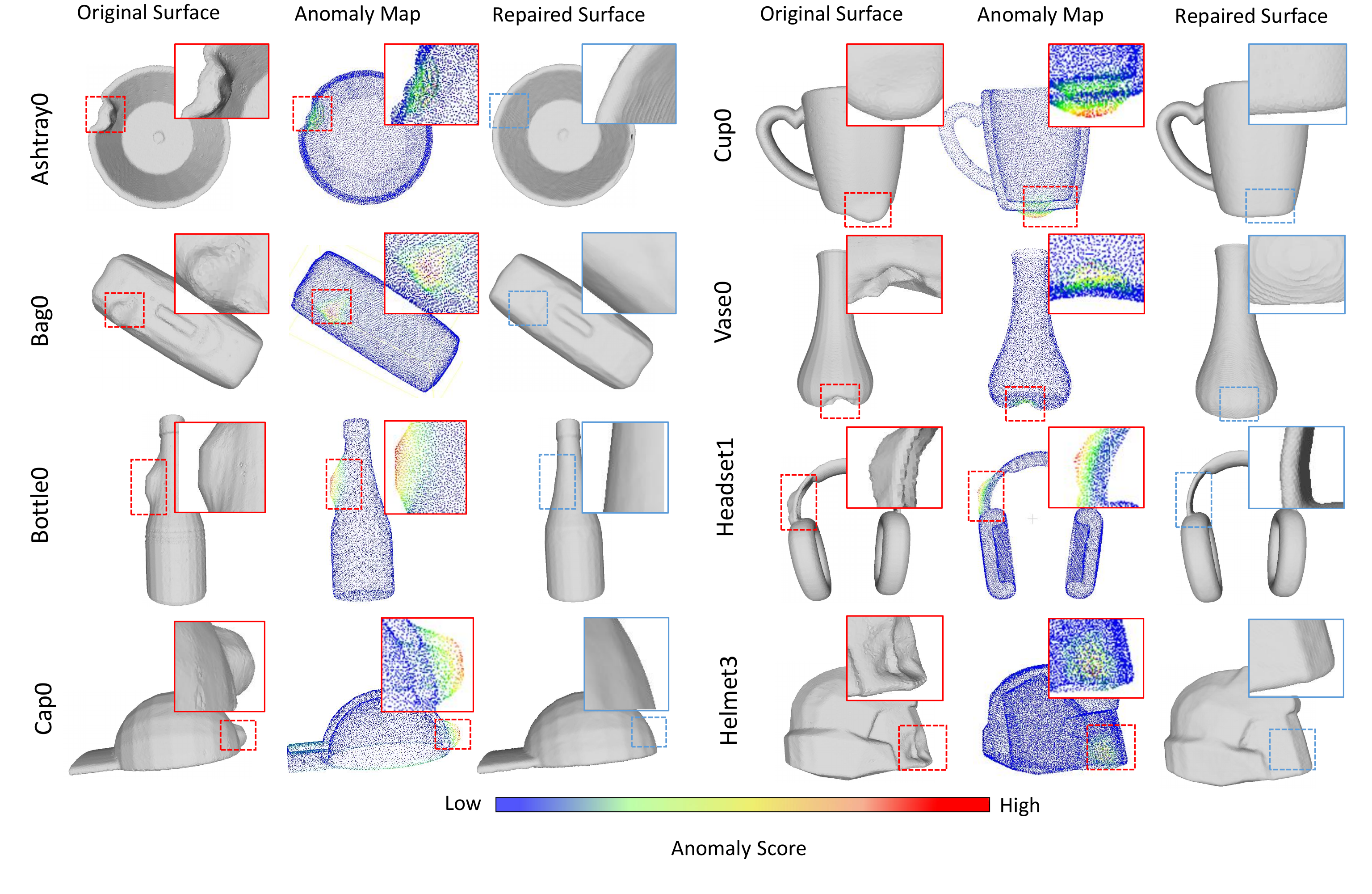}\vspace{-3mm}
  \caption{\textcolor{black}{{Qualitative anomaly localization results on Anomaly-ShapeNet.}
}}
  \label{fig:repair_suppl0}\vspace{5mm}

  \centering
  \setlength{\abovecaptionskip}{0.1cm}
  \includegraphics[width=1\textwidth]{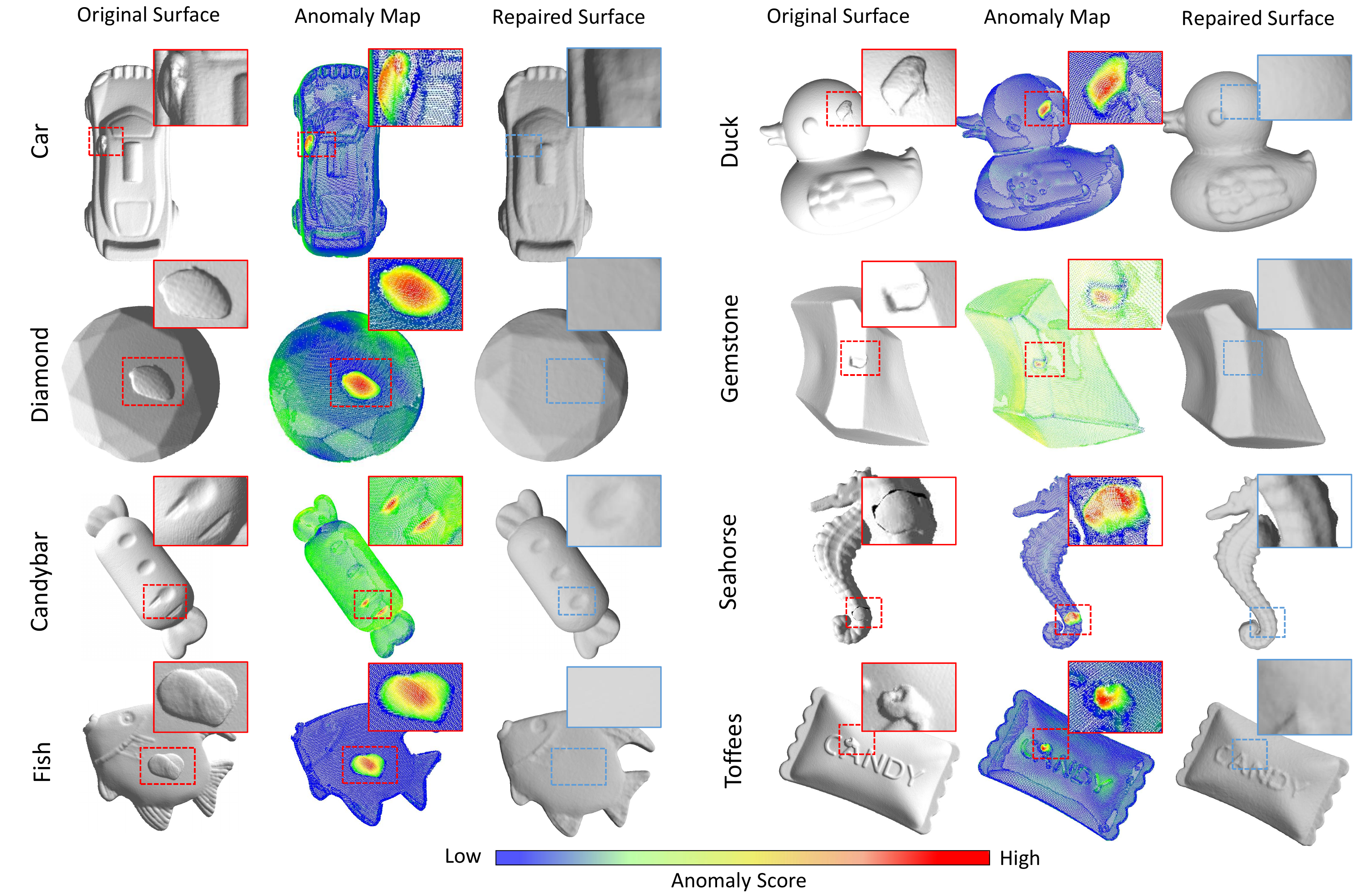}
  \caption{\textcolor{black}{{Qualitative anomaly localization results on Real3D-AD.}
}}
  \label{fig:repair_suppl1}\vspace{-3mm}
\end{figure*}

Due to page limit, only a limited number of qualitative results are presented in the main text. To offer a more comprehensive and intuitive visualization of our results, we provide additional qualitative results in Figure~\ref{fig:repair_suppl0}, \ref{fig:repair_suppl1}. Specifically, the first column displays the original surface, the second column shows the anomaly map, and the third column presents the repaired surface.




\begin{table*}[th]
  \centering    \renewcommand{\arraystretch}{1.10}\setlength\tabcolsep{0.8mm}     \setlength{\abovecaptionskip}{0.1cm}
  \begin{adjustbox}{max width=\textwidth}
    \begin{tabular}{l|*{27}{S[table-format=1.3]}}  
    \toprule

    Method & \multicolumn{1}{c}{cap0} & \multicolumn{1}{c}{cap3} & \multicolumn{1}{c}{helmet3} & \multicolumn{1}{c}{cup0} & \multicolumn{1}{c}{bowl4} & \multicolumn{1}{c}{vase3} & \multicolumn{1}{c}{headset1} & \multicolumn{1}{c}{eraser0} & \multicolumn{1}{c}{vase8} & \multicolumn{1}{c}{cap4} & \multicolumn{1}{c}{vase2} & \multicolumn{1}{c}{vase4} & \multicolumn{1}{c}{helmet0} & \multicolumn{1}{c}{bucket1} & \multicolumn{1}{c}{vase7} & \multicolumn{1}{c}{helmet2} & \multicolumn{1}{c}{cap5} & \multicolumn{1}{c}{shelf0} & \multicolumn{1}{c}{bowl5} & \multicolumn{1}{c}{bowl3} & \multicolumn{1}{c}{helmet1} \\
    \midrule
 
   \rowcolor[HTML]{F1F3F4}   BTF(FPFH) & 0.618 & 0.522 & 0.444 & 0.586 & 0.609 & \underline{0.699} & 0.490 & 0.719 & 0.668 & 0.520 & 0.546 & 0.510 & 0.571 & 0.633 & 0.518 & 0.542 & 0.586 & 0.609 & 0.699 & 0.490 & 0.719 \\
       BTF(FPFH) with PAM &  0.741 & \underline{0.740} & 0.567 & 0.881 & 0.644 & 0.433 & 0.680 & 0.810 & 0.558 & 0.533 & 0.652 & 0.609 & \underline{0.635} & 0.444 & 0.629 & 0.684 & 0.653 & 0.530 & 0.632 & \underline{0.786} & 0.648 \\

  \rowcolor[HTML]{F1F3F4}    Patchcore(FPFH)& 0.580 & 0.453 & 0.404 & 0.600 & 0.494 & 0.449 & 0.637 & 0.657 & 0.662 & \textbf{0.757} & 0.721 & 0.506 & 0.546 & 0.551 & 0.693 & 0.425 & 0.790 & 0.494 & 0.558 & 0.537 & 0.484 \\
    Patchcore(FPFH) with PAM& \textbf{0.996} & \textbf{0.811} & \underline{0.624} & \underline{0.914} & \underline{0.863} & 0.679 & \underline{0.781} & \textbf{1.000} & \underline{0.852} & \underline{0.744} & \textbf{1.000} & \underline{0.848} & 0.432 & 0.603 & \underline{0.881} & \textbf{0.788} & \underline{0.800} & \textbf{0.786} & \underline{0.828} & 0.644 & \underline{0.729} \\
  
    \rowcolor[HTML]{F1F3F4} Patchcore(PointMAE)& 0.589 & 0.476 & 0.424 & 0.610 & 0.501 & 0.460 & 0.627 & 0.677 & 0.663 & 0.727 & \underline{0.741} & 0.516 & 0.556 & 0.561 & 0.650 & 0.447 & 0.538 & 0.523 & 0.593 & 0.579 & 0.552 \\
    Patchcore(PointMAE) with PAM& 0.619 & 0.660 & 0.455 & 0.605 & 0.385 & 0.664 & 0.462 & 0.790 & 0.694 & 0.407 & 0.614 & 0.542 & 0.441 & \textbf{0.832} & 0.600 & 0.609 & 0.744 & 0.580 & 0.554 & 0.470 & 0.338 \\  \midrule
  \rowcolor[HTML]{F1F3F4}    \textbf{PASDF(Ours)} & \underline{0.852} & 0.649 & \textbf{0.846} & \textbf{0.971} & \textbf{0.933} & \textbf{0.806} & \textbf{0.795} & \underline{0.952} & \textbf{0.924} & 0.646 & \textbf{1.000} & \textbf{0.912} & \textbf{0.812} & \underline{0.775} & \textbf{1.000} & \underline{0.765} & \textbf{0.853} & \underline{0.713} & \textbf{0.912} & \textbf{1.000} & \textbf{0.938} \\
    \midrule \midrule
    Method & \multicolumn{1}{c}{bottle3} & \multicolumn{1}{c}{vase0} & \multicolumn{1}{c}{bottle0} & \multicolumn{1}{c}{tap1} & \multicolumn{1}{c}{bowl0} & \multicolumn{1}{c}{bucket0} & \multicolumn{1}{c}{vase5} & \multicolumn{1}{c}{vase1} & \multicolumn{1}{c}{vase9} & \multicolumn{1}{c}{ashtray0} & \multicolumn{1}{c}{bottle1} & \multicolumn{1}{c}{tap0} & \multicolumn{1}{c}{phone} & \multicolumn{1}{c}{cup1} & \multicolumn{1}{c}{bowl1} & \multicolumn{1}{c}{headset0} & \multicolumn{1}{c}{bag0} & \multicolumn{1}{c}{bowl2} & \multicolumn{1}{c|}{jar0} & \multicolumn{2}{|c}{\textbf{Mean}} \\
    \midrule

   \rowcolor[HTML]{F1F3F4}   BTF(FPFH) & 0.322 & 0.342 & 0.344 & 0.546 & 0.509 & 0.401 & 0.409 & 0.219 & 0.268 & 0.420 & 0.546 & 0.560 & 0.671 & 0.610 & 0.668 & 0.520 & 0.546 & 0.510 & 0.424 & \multicolumn{2}{|c}{0.528} \\
       BTF(FPFH) with PAM & 0.822 & 0.779 & 0.705 & 0.533 & 0.833 & 0.724 & 0.495 & 0.457 & 0.612 & 0.686 & 0.523 & 0.661 & 0.676 & 0.552 & 0.552 & 0.693 & \underline{0.714} & 0.559 & 0.295 &  \multicolumn{2}{|c}{0.579 } \\
   
    \rowcolor[HTML]{F1F3F4}  Patchcore(FPFH) & 0.572 & 0.455 & 0.604 & \underline{0.766} & 0.504 & 0.469 & 0.417 & 0.423 & 0.660 & 0.587 & 0.667 & \underline{0.753} & 0.388 & 0.586 & 0.639 & 0.583 & 0.571 & \underline{0.615} & 0.472 & \multicolumn{2}{|c}{0.568} \\
     Patchcore(FPFH) with PAM & \underline{0.990} & \underline{0.925} & \underline{0.962} & 0.507 & \underline{0.974} & \textbf{0.981} & \underline{0.752} & \underline{0.767} & \textbf{0.882} & \underline{0.919} & \underline{0.951} & 0.691 & \textbf{1.000} & \underline{0.771} & \underline{0.819} & \underline{0.942} & 0.710 & 0.407 & \textbf{1.000} &  \multicolumn{2}{|c}{\underline{0.814}}  \\

    \rowcolor[HTML]{F1F3F4} Patchcore(PointMAE) & 0.650 & 0.447 & 0.513 & 0.538 & 0.523 & 0.593 & 0.579 & 0.552 & 0.629 & 0.591 & 0.601 & 0.458 & 0.488 & 0.556 & 0.629 & 0.591 & 0.601 & 0.458 & 0.483 & \multicolumn{2}{|c}{0.562} \\

    Patchcore(PointMAE) with PAM & 0.517 & 0.887 & 0.652 & 0.537 & 0.641 & 0.559 & 0.657 & 0.443 & 0.564 & 0.781 & 0.779 & 0.736 & \underline{0.710} & 0.481 & 0.522 & 0.707 & 0.538 & 0.563 & \underline{0.738} &   \multicolumn{2}{|c}{0.627} \\ \midrule

  \rowcolor[HTML]{F1F3F4}    \textbf{PASDF(Ours)} & \textbf{1.000} & \textbf{1.000} & \textbf{1.000} & \textbf{0.793} & \textbf{1.000} & \underline{0.968} & \textbf{1.000} & \textbf{0.929} & \underline{0.836} & \textbf{1.000} & \textbf{1.000} & \textbf{0.882} & \textbf{1.000} & \textbf{0.857} & \textbf{0.948} & \textbf{1.000} & \textbf{0.995} & \textbf{1.000} & \textbf{1.000} & \multicolumn{2}{|c}{\textbf{0.900}} \\
    \bottomrule
    \end{tabular}%
  \end{adjustbox}
  \caption{O-AUROC score~($\uparrow$) on Anomaly-ShapeNet dataset. The best and second-best results are marked in \textbf{bold} and \underline{underlined}.
  }
  \label{O-AUROC-PAM}\vspace{-2mm}
\end{table*}

\begin{table*}[th]
  \centering      \renewcommand{\arraystretch}{1.10}\setlength{\abovecaptionskip}{0.1cm}
  \setlength\tabcolsep{0.8mm}
  \begin{adjustbox}{max width=\textwidth}
    \begin{tabular}{l|*{27}{S[table-format=1.3]}}
    \toprule
Method & \multicolumn{1}{c}{cap0} & \multicolumn{1}{c}{cap3} & \multicolumn{1}{c}{helmet3} & \multicolumn{1}{c}{cup0} & \multicolumn{1}{c}{bowl4} & \multicolumn{1}{c}{vase3} & \multicolumn{1}{c}{headset1} & \multicolumn{1}{c}{eraser0} & \multicolumn{1}{c}{vase8} & \multicolumn{1}{c}{cap4} & \multicolumn{1}{c}{vase2} & \multicolumn{1}{c}{vase4} & \multicolumn{1}{c}{helmet0} & \multicolumn{1}{c}{bucket1} & \multicolumn{1}{c}{vase7} & \multicolumn{1}{c}{helmet2} & \multicolumn{1}{c}{cap5} & \multicolumn{1}{c}{shelf0} & \multicolumn{1}{c}{bowl5} & \multicolumn{1}{c}{bowl3} & \multicolumn{1}{c}{helmet1} \\
    \midrule

   \rowcolor[HTML]{F1F3F4}   BTF(FPFH) & 0.730 & 0.658 & 0.724 & 0.790 & 0.679 & 0.699 & 0.591 & 0.719 & 0.662 & 0.524 & 0.646 & 0.710 & 0.575 & 0.633 & 0.540 & 0.643 & 0.586 & 0.619 & 0.699 & 0.690 & \textbf{0.749} \\
    BTF(FPFH) with PAM &  0.864 & 0.699 & 0.608 & 0.892 & 0.611 & 0.568 & 0.505 & 0.894 & 0.586 & 0.689 & 0.872 & 0.650 & 0.731 & 0.648 & 0.682 & 0.524 & 0.803 & 0.769 & 0.715 & 0.807 & 0.533\\

     \rowcolor[HTML]{F1F3F4}  Patchcore(FPFH) & 0.472 & 0.653 & 0.737 & 0.655 & 0.720 & 0.430 & 0.464 & 0.810 & 0.575 & 0.595 & 0.721 & 0.505 & 0.548 & 0.571 & 0.693 & 0.455 & 0.795 & 0.613 & 0.358 & 0.327 & 0.489 \\

       Patchcore(FPFH) with PAM & \textbf{0.973} & \textbf{0.921} & \underline{0.856} & \textbf{0.976} & \underline{0.860} & \underline{0.752} & \underline{0.850} & \textbf{0.946} & \underline{0.856} & \textbf{0.966} & \textbf{0.978} & \textbf{0.964} & \underline{0.806} & 0.783 & \underline{0.958} & \textbf{0.818} & \textbf{0.958} & \textbf{0.883} & \underline{0.774} & \textbf{0.972} & \underline{0.732}  \\
  \rowcolor[HTML]{F1F3F4}    Patchcore(PointMAE) & 0.544 & 0.488 & 0.615 & 0.510 & 0.501 & 0.465 & 0.423 & 0.378 & 0.364 & 0.725 & 0.742 & 0.523 & 0.580 & 0.754 & 0.651 & 0.651 & 0.545 & 0.543 & 0.562 & 0.581 & 0.562 \\
Patchcore(PointMAE) with PAM & 0.739 & 0.752 & 0.660 & 0.663 & 0.436 & 0.601 & 0.605 & 0.626 & 0.680 & 0.702 & 0.691 & 0.748 & 0.569 & \textbf{0.880} & 0.568 & 0.642 & 0.802 & 0.687 & 0.528 & 0.586 & 0.572\\ \midrule

  \rowcolor[HTML]{F1F3F4}    \textbf{PASDF(Ours)} & \underline{0.948} & \underline{0.861} & \textbf{0.958} & \underline{0.948} & \textbf{0.865} & \textbf{0.868} & \textbf{0.891} & \underline{0.945} & \textbf{0.909} & \underline{0.894} & \underline{0.956} & \underline{0.899} & \textbf{0.816} & \underline{0.824} & \textbf{0.959} & \underline{0.809} & \underline{0.920} & \underline{0.865} & \textbf{0.909} & \underline{0.939} & 0.646 \\
    \midrule \midrule
Method & \multicolumn{1}{c}{bottle3} & \multicolumn{1}{c}{vase0} & \multicolumn{1}{c}{bottle0} & \multicolumn{1}{c}{tap1} & \multicolumn{1}{c}{bowl0} & \multicolumn{1}{c}{bucket0} & \multicolumn{1}{c}{vase5} & \multicolumn{1}{c}{vase1} & \multicolumn{1}{c}{vase9} & \multicolumn{1}{c}{ashtray0} & \multicolumn{1}{c}{bottle1} & \multicolumn{1}{c}{tap0} & \multicolumn{1}{c}{phone} & \multicolumn{1}{c}{cup1} & \multicolumn{1}{c}{bowl1} & \multicolumn{1}{c}{headset0} & \multicolumn{1}{c}{bag0} & \multicolumn{1}{c}{bowl2} & \multicolumn{1}{c|}{jar0} & \multicolumn{2}{|c}{\textbf{Mean}} \\ 
    \midrule

    \rowcolor[HTML]{F1F3F4} BTF(FPFH) & 0.622 & 0.642 & 0.641 & 0.596 & 0.710 & 0.401 & 0.429 & 0.619 & 0.568 & 0.624 & 0.549 & 0.568 & 0.675 & 0.619 & \underline{0.768} & 0.620 & 0.746 & 0.518 & 0.427 & \multicolumn{2}{|c}{0.628} \\
    BTF(FPFH) with PAM & 0.809 & 0.837 & 0.880 & 0.674 & 0.769 & 0.772 & 0.567 & 0.652 & 0.695 & 0.693 & 0.696 & 0.647 & 0.869 & 0.687 & 0.548 & 0.717 & 0.841 & 0.621 & 0.852 &  \multicolumn{2}{|c}{0.683}\\

    \rowcolor[HTML]{F1F3F4} Patchcore(FPFH) & 0.512 & 0.655 & 0.654 & \underline{0.768} & 0.524 & 0.459 & 0.447 & 0.453 & 0.663 & 0.597 & 0.687 & 0.733 & 0.488 & 0.596 & 0.531 & 0.583 & 0.574 & 0.625 & 0.478 & \multicolumn{2}{|c}{0.580} \\

   Patchcore(FPFH) with PAM & \underline{0.889} & \underline{0.929} & \textbf{0.985} & 0.716 & \textbf{0.981} & \underline{0.842} & \underline{0.713} & \underline{0.757} & \textbf{0.905} & \underline{0.774} & \underline{0.841} & 0.827 & \underline{0.943} & 0.758 & 0.732 & \underline{0.807} & \underline{0.895} & \underline{0.814} & \textbf{0.985} & \multicolumn{2}{|c}{\underline{0.867}} \\
   \rowcolor[HTML]{F1F3F4}   Patchcore(PointMAE) & 0.653 & 0.677 & 0.553 & 0.541 & 0.527 & 0.586 & 0.572 & 0.551 & 0.423 & 0.495 & 0.606 & \underline{0.858} & 0.886 & \underline{0.856} & 0.524 & 0.575 & 0.674 & 0.515 & 0.487 & \multicolumn{2}{|c}{0.577} \\
        Patchcore(PointMAE) with PAM & 0.883 & 0.835 & 0.691 & 0.681 & 0.656 & 0.574 & 0.602 & 0.675 & 0.719 & 0.765 & 0.735 & 0.687 & 0.792 & 0.569 & 0.566 & 0.646 & 0.593 & 0.525 & 0.774 & \multicolumn{2}{|c}{0.681 } \\
 \midrule
  \rowcolor[HTML]{F1F3F4}    \textbf{PASDF(Ours)} & \textbf{0.948} & \textbf{0.944} & \underline{0.951} & \textbf{0.902} & \underline{0.963} & \textbf{0.875} & \textbf{0.915} & \textbf{0.797} & \underline{0.863} & \textbf{0.919} & \textbf{0.926} & \textbf{0.884} & \textbf{0.951} & \textbf{0.884} & \textbf{0.900} & \textbf{0.863} & \textbf{0.958} & \textbf{0.816} & \underline{0.959} & \multicolumn{2}{|c}{\textbf{0.897}} \\
    \bottomrule
    \end{tabular}%
  \end{adjustbox}
  \caption{P-AUROC score~($\uparrow$) on Anomaly-ShapeNet dataset. The best and second-best results are marked in \textbf{bold} and \underline{underlined}.
  }
  \label{P-AUROC-PAM}\vspace{-3mm}
\end{table*}

\section{Detailed Results of Ablation Studies}
In the main text (Table 3), we present the overall results of the PAM ablation study, demonstrating its effectiveness in improving anomaly detection performance across different baseline models on Anomaly-ShapeNet. To provide a more detailed analysis, we report the per-class quantitative results in Table~\ref{O-AUROC-PAM},~\ref{P-AUROC-PAM}. These results offer a finer-grained evaluation of PAM’s impact on individual object categories, further validating its robustness and generalization ability.

\end{document}